\pdfoutput=1
\documentclass{article}

\usepackage{arxiv_layout}

\usepackage[utf8]{inputenc}
\usepackage{microtype}
\usepackage{amsmath}
\usepackage{amssymb}
\usepackage{booktabs}
\usepackage{array}
\usepackage{float}
\usepackage{graphicx}
\usepackage{xcolor}
\usepackage{tcolorbox}
\usepackage{caption}
\usepackage{url}
\graphicspath{{figures/}}
\usepackage[colorlinks=true, linkcolor=blue, citecolor=blue, urlcolor=blue]{hyperref}

\title{The Alignment Tax Is Graded:\\
Task-Selective Capability Loss Under\\
Length-Normalized DPO with SFT Retention}

\author{%
  Ye Yuan\thanks{Ph.D. in Physics, University College Dublin.}\\
  Independent Researcher\\
  \texttt{ye.yuan.phd@gmail.com}%
}

\begin{document}
\maketitle

\begin{abstract}
\textbf{Setting.} This is an empirical characterization study; we propose no new method and run no comparisons against alternative objectives. We fine-tune Qwen3.5-4B with sparse LoRA and length-normalized DPO plus an SFT retention term (the RPO objective) on structured financial-filing tasks, and track downstream capability as the implicit reward margin grows to ${\sim}7$ $\beta$-scaled units (nf4 training; ${\sim}3$ under fp16 measurement on the unquantized base). \textbf{Question.} Does margin growth degrade capability uniformly across tasks, or selectively? \textbf{Findings.} In a pre-registered study (8 seeds, 6 unseen at freeze) with artifact-aware scoring and TOST equivalence testing, the alignment tax is graded and task-selective: extraction is preserved ($\Delta = -0.55$ pts, equivalent at $\pm5$ pts), json loses little ($\Delta = -3.52$ pts, equivalence fails on a clustered item-level interval), and router degrades severely ($\Delta = -30$ pts). Router's loss concentrates in a margin 1--3 damage window containing a transient format-break spike. The margin is accounted for almost entirely (${\geq}96\%$) by suppression of rejected completions, with chosen reward approximately flat, consistent with the retention term anchoring it; we do not attribute the behavioral damage to any single gradient channel. Gold-logp deflation is task-selective (extraction $-0.15$, json $-1.04$, router $-1.47$ nats/token under fp16), with gold selectively protected relative to incorrect controls on extraction (8/8 seeds). QLoRA reward margins differ up to $3.4\times$ between nf4 training and unquantized fp16 measurement, with an axis-dependent gap. \textbf{Scope.} Length-normalized rewards with SFT retention, sparse LoRA on 8/32 layers, one 4B model, easily-separable synthetic preferences, three structured tasks. A no-retention control and a measured format-alignment metric are left to future work.
\end{abstract}

\section{Introduction}

DPO's implicit reward margin grows large under length-normalized rewards, reaching ${\sim}7$ $\beta$-scaled units in our experiments. This echoes the unbounded growth IPO was designed to prevent (our margins are per-token; see \S\ref{sec:setup} for the conversion). Several lines of work treat this growth as a potential concern. Razin et al.\ \cite{razin2025displacement} identify likelihood displacement, where chosen log-prob decreases and probability mass may flow to out-of-distribution completions. IPO \cite{azar2024ipo} bounds the margin, treating unbounded growth as the core problem. Cal-DPO \cite{xiao2024caldpo} observes absolute likelihood deflation and proposes a calibration fix. These analyses draw mainly on judge-based evaluation or full fine-tuning at large scale; what margin growth does to downstream capability in the common practical regime (small model, LoRA, length-normalized rewards, structured exact-format tasks) has not been documented. This is an empirical characterization study: we do not propose a new training algorithm, but measure how an existing one (length-normalized DPO with an SFT regularization term) affects downstream capability as its implicit reward margin grows. A methodological gap compounds the empirical one: we are not aware of prior DPO work that uses pre-registered TOST as its primary capability test. Studies sometimes report ``no significant difference'' without the power to distinguish ``preserved'' from ``underpowered to detect a real drop.''

We present a pre-registered study (8 seeds, 6 unseen at freeze) of a 4B model fine-tuned with LoRA. We measure behavioral capability (free-generation with artifact-aware scoring) and probability-layer dynamics (per-token gold/control log-probabilities) as the margin grows to ${\sim}7$. That value is from the nf4 training backend; the margin is ${\sim}3$ under fp16 measurement on the unquantized base. The objective we study combines length-normalized DPO with an SFT retention term (weight 0.5) that anchors the likelihood of the chosen completion. This is a common practical choice. As we show, almost the entire margin (${\geq}96\%$) is then accounted for by suppression of rejected completions.

The alignment tax is real but graded. It is undetectable on extraction ($\Delta = -0.55$, equivalent at $\pm5$), small but statistically genuine on json ($\Delta = -3.52$, equivalence fails on the item-level interval), and catastrophic on router ($\Delta = -30$). The degradation ordering tracks format-alignment to the preference data's dominant style axis: router is worst on 8/8 seeds; extraction loses less than json on 6/8 seeds; the full ordering holds in the mean. With three tasks, we report this co-occurrence without claiming causation.

The mechanism is asymmetric. Chosen reward stays approximately stationary, consistent with the retention term anchoring it, while DPO gradients drive rejected reward to $-14.2$ (nf4). As a result, margin growth comes almost entirely (${\geq}96\%$) from rejected-completion collapse. The style axis dominates this suppression in aggregate under both backends, but per-pair ordering is backend-dependent (\S\ref{sec:mechanism}). A pre-registered prediction about which preference axis dominates per-pair margin was falsified (\S\ref{sec:mechanism}). Gold-logp deflation is itself task-selective and directionally consistent with behavioral outcomes, though non-linearly related (\S\ref{sec:mechanism}--\ref{sec:gold}). A naive strict-match scorer masked this picture by producing a false capability collapse (see the scorer artifact in \S\ref{sec:tasks}). Artifact-aware evaluation and pre-registered TOST equivalence testing prevented that error. Here TOST serves as the primary evaluation framework for capability assessment.

Our contributions are three, all empirical or methodological; none is a new training method. (1)~A graded, task-selective alignment tax, with non-monotonic format-break dynamics at margin ${\approx}3$ and a falsified pre-registered prediction on held-out per-pair margin (\S\ref{sec:graded}--\ref{sec:mechanism}). Margin growth is accounted for almost entirely (${\geq}96\%$) by suppression of rejected completions. The source of the behavioral damage is not attributed pending a no-retention control, and the format-alignment hypothesis for which tasks are affected is consistent with the data but not directly measured. (2)~Task-selective gold-logp deflation, with selective protection demonstrated on extraction (8/8 seeds, paired statistic, fp16) and the same task ordering under both backends, though the nf4 comparison also changes the baseline checkpoint (\S\ref{sec:gold}). (3)~An evaluation protocol that applies established tools (artifact-aware answer canonicalization and pre-registered TOST equivalence testing) to DPO capability assessment (\S\ref{sec:method}). Its falsification chronology (three falsified hypotheses, two refined characterizations) documents how pre-registration forced successive refinement of initially simpler claims (\S\ref{sec:chronology}).

This study operates within explicit scope conditions: sparse LoRA on 8/32 layers (0.034\% of parameters; not full fine-tuning), a single 4B model, length-normalized rewards with an SFT retention term (weight 0.5), easily-separable synthetic preferences, and three structured exact-format tasks. We do not study safety or refusal tasks, where Razin et al.\ demonstrated catastrophic displacement. We also do not study full fine-tuning, naturalistic human-written preferences, or models above 4B parameters.

\section{Related Work}

We do not run any of the methods below; each paragraph states how our observations, in our specific regime, relate to the cited claims.

\textbf{Likelihood displacement.} Razin et al.\ (ICLR 2025) \cite{razin2025displacement} identify likelihood displacement under DPO: chosen log-prob decreases while mass may flow to out-of-distribution or harmful completions. They introduce the CHES score to flag pairs whose embedding proximity drives catastrophic displacement. Their analysis notes displacement is ``not necessarily problematic''; it is catastrophic only when mass flows to opposite-meaning responses. The objective we study includes an SFT retention term that anchors chosen likelihood. It is equivalent to RPO (Liu et al.\ \citep{liu2024rpo}) and is the mitigation Razin et al.\ themselves propose. This chosen-decline dynamic is therefore largely suppressed (\S\ref{sec:mechanism}); whether displacement would emerge without it is untested. Pipano et al.\ \citep{pipano2026displacement} characterize which $f$-divergence losses are displacement-resistant by construction. The objective we study instead retains the standard DPO loss with SFT anchoring.

\textbf{The squeezing effect.} Ren and Sutherland (ICLR 2025) \cite{ren2025squeezing} show DPO's gradient is dominated by negative gradients on rejected completions, concentrating mass onto the highest-probability token per position. They demonstrate this at the single-token level on Pythia and Qwen. We observe consistent dynamics on multi-token structured completions, though the transfer is assumed rather than proven. Our reward decomposition (\S\ref{sec:mechanism}) is consistent with squeezing (rejected reward $-14.2$, chosen reward within $-0.072$ to $+0.052$, nf4). It also offers a candidate explanation for task-selective gold deflation (\S\ref{sec:gold}). Extraction outputs share entities with training data and may gain from mass concentration onto gold tokens. Router's terse format is maximally dissimilar from the verbose pairs, so router may not.

\textbf{Ordering preservation and calibration.} Cal-DPO (Xiao et al., NeurIPS 2024) \cite{xiao2024caldpo} observes DPO preserves preference ordering while absolute likelihoods deflate, and calibrates the implicit reward to prevent chosen-reward decline. The retention term in the objective we study achieves similar anchoring (\S\ref{sec:setup}). The flatness of chosen reward does not show that Cal-DPO's concern is unfounded; it is consistent with the retention term addressing the concern. Even with chosen anchored, the behavioral tax remains on off-target tasks (\S\ref{sec:graded}). The margin is accounted for by rejected suppression (\S\ref{sec:mechanism}), but the source of the damage is not attributed.

\textbf{Bounding the margin.} IPO (Azar et al., AISTATS 2024) \cite{azar2024ipo} replaces DPO's log-sigmoid with a squared loss to bound the margin, treating unbounded growth as the core pathology. Our margin grows to ${\sim}7$ with graded, format-selective effects: the damage is task-selective rather than uniform (\S\ref{sec:graded}--\ref{sec:nonmono}). We did not train with IPO; in our regime, margin growth to ${\sim}7$ co-occurs with damage that is task-selective and concentrated in a margin 1--3 window.

\textbf{Chosen-reward predictions.} Rafailov et al.\ (COLM 2024) \cite{rafailov2024qfunction} predict from a Q-function framework that chosen log-prob should decrease under DPO. The retention term in the objective we study largely counteracts this at the endpoint, though measurement on the unquantized base reveals a residual transient dip (\S\ref{sec:priorwork}).

\textbf{Equivalence testing and scoring.} Manzoni (2026) \cite{manzoni2026tost} uses TOST with 8 seeds for SFT capability preservation in MoE fine-tuning. A concurrent paper (Bao and Abisado, \emph{Symmetry} 2026) \cite{bao2026pareto} applies TOST as incidental validation in a DPO context. We are not aware of prior DPO work that uses pre-registered TOST as its primary capability test. TOST itself is well-established in bioequivalence. Baxi (2025) \cite{baxi2025scoring} separates constraint compliance from semantic accuracy, establishing the decomposition. What we add is the trajectory: a transient format-break spike versus a monotonic content-error rise (\S\ref{sec:nonmono}). Minerva (Lewkowycz et al., 2022) \cite{lewkowycz2022minerva} provides precedent for answer normalization preventing exact-match false negatives. Our canonicalization result is a concrete alignment-specific instance (see the scorer artifact in \S\ref{sec:tasks}).

\textbf{DPO dynamics and alignment-tax context.} Li (2026) \cite{li2026oxrl} finds none of 20 DPO variants outperforms vanilla DPO under Bonferroni correction. Young (2026) \cite{young2026geometry} frames the alignment tax as the projection of the safety direction onto the capability subspace. Our format-alignment observation (\S\ref{sec:graded}) may be read as an empirical instance of this framing, though it is untested beyond three tasks. Pollanen (2026) \cite{pollanen2026margin} reports margin can anticorrelate with reasoning capability ($r = -0.91$, LLaMA on logic tasks, sweeping $\beta$). They vary $\beta$ rather than observing within-run growth, so the result is related to, but distinct from, our trajectory analysis.

\section{Method}
\label{sec:method}

\subsection{Setup}
\label{sec:setup}
We fine-tune Qwen3.5-4B with LoRA ($r = 8$, $\alpha = 16$, dropout $= 0.05$) using vanilla length-normalized DPO ($\beta = 0.5$) with an SFT retention term. Let $\pi_\theta$ denote the policy and $\pi_{\text{ref}}$ the SFT reference. For a prompt $x$ and completion $y$ of length $|y|$ tokens, the length-normalized log-probability is
\begin{equation}
\ell_\theta(y \mid x) = \frac{1}{|y|} \sum_{t=1}^{|y|} \log \pi_\theta(y_t \mid x, y_{<t}).
\label{eq:lnlogp}
\end{equation}
Here $|y|$ counts completion tokens only; the same completion-only masking applies to $\ell$ in Eqs.~\eqref{eq:pref} and~\eqref{eq:total}. Length normalization follows the per-token formulation discussed by Meng et al.\ \citep{meng2024simpo}. The implicit reward of a completion is $r_\theta(x, y) = \ell_\theta(y \mid x) - \ell_{\text{ref}}(y \mid x)$, the policy log-probability minus the reference log-probability. Given a preference pair $(x, y_w, y_l)$ with chosen $y_w$ and rejected $y_l$, the $\beta$-scaled preference margin and the DPO preference loss are
\begin{equation}
m = \beta\,\big[r_\theta(x, y_w) - r_\theta(x, y_l)\big], \qquad \mathcal{L}_{\text{pref}} = -\log \sigma(m), \qquad \beta = 0.5.
\label{eq:pref}
\end{equation}
Here $\sigma$ is the logistic sigmoid. Because Eq.~\eqref{eq:lnlogp} is per-token and $\beta = 0.5$, a reported margin of 7 corresponds to a per-token log-ratio difference of 14 nats. Relative to the reference, the policy's per-token probability on rejected completions falls by a geometric-mean factor of about $e^{14} \approx 1.2 \times 10^{6}$, while on chosen completions it stays near the reference (\S\ref{sec:mechanism}). This is near-total suppression of the rejected text. Displacement and IPO analyses typically reason about sequence-level log-ratios, so our margins are not directly comparable in scale.

The total training objective adds an assistant-only cross-entropy on the chosen completion, with system and user tokens masked:
\begin{equation}
\mathcal{L} = \mathcal{L}_{\text{pref}} + \lambda\, \mathcal{L}_{\text{sft}}, \qquad \mathcal{L}_{\text{sft}} = -\ell_\theta(y_w \mid x), \qquad \lambda = 0.5.
\label{eq:total}
\end{equation}
Equation~\eqref{eq:total} is the Regularized Preference Optimization (RPO) objective of Liu et al.\ \citep{liu2024rpo}; a related DPO+NLL objective appears in Pang et al.\ \citep{pang2024iterative}. We refer to the $\lambda \mathcal{L}_{\text{sft}}$ term as the SFT retention term throughout. Razin et al.\ \citep{razin2025displacement} identify this SFT regularization as a mitigation for likelihood displacement. The retention term anchors the likelihood of the chosen completion during preference optimization.

Qwen3.5-4B is a hybrid architecture with 24 linear-attention layers and 8 full-attention layers (every 4th: layers 3, 7, 11, 15, 19, 23, 27, 31). LoRA adapts only the 8 full-attention layers (q/k/v/o\_proj; 1.57M trainable parameters, 0.034\% of 4.66B total), leaving the 24 linear-attention layers and all MLPs untouched. Training uses AdamW (lr $5\times10^{-5}$, constant schedule, weight decay 0.01), gradient clipping 1.0, batch size 8, and 180 fixed optimizer steps with no loss-based early stop. A divergence guard aborts on NaN or smoothed margin $>20$. The confirmatory run (8 seeds; seeds 0--7) starts from the same SFT checkpoint (hash-verified; Appendix~\ref{app:config}). The SFT checkpoint is a QLoRA adapter with the same LoRA configuration, trained for 4 epochs on 862 GPT-4o-mini-generated chat examples in the same SEC-filing domain (analytical prose answers, JSON tool calls, router labels, and answer-quality labels; none in the extraction output format). DPO continues training this SFT adapter rather than stacking a new adapter or merging, and the frozen SFT adapter is the reference $\pi_{\text{ref}}$. The chat template is byte-identical across sources, and tokenization is identical on all study inputs. The SFT adapter, DPO adapters, and base tokenizer configs differ in metadata (added-token count, pre-tokenizer regex) but produce identical token IDs on all 6{,}024 evaluation and preference strings, verified by direct token-ID comparison.

The preference data is 647 pairs (549 train, 98 held-out; deterministic split) on two axes. The correctness axis has 200 tool\_calling pairs, 171 of them in the 549-pair training split. Its rejected completions are wrong-tool swaps: identical arguments, randomly substituted tool name, length-matched at ${\sim}7.5$ words. The style axis has 447 pairs, 378 of them in the training split. Its chosen completions are verbose analytical responses ($\sim$190--226 words). Its rejected completions are GPT-4o-mini degradations with numbers, dates, and section references stripped and text truncated to $\sim$86--100 words. Because the rejected completions strip numerical values, dates, and section references, the style axis confounds presentation style with factual content. Length normalization is applied because otherwise the 447 style pairs would be winnable on length alone.

\subsection{Tasks and Evaluation Suite}
\label{sec:tasks}
We evaluate three structured exact-format tasks, 160 items each, on SEC-filing queries (Table~\ref{tab:tasks}). Evaluation prompts were generated by GPT-4o-mini from targets specified in code (a router category; a tool and ticker; or a ticker, section, and filing date), so each gold label is fixed by construction. Extraction passages are therefore synthetic filing-style text, not verbatim EDGAR excerpts. Router gold labels were independently checked by GPT-4o; items it disagreed with or flagged as ambiguous were replaced, and all 160 final router items carry agreed labels. Free generation uses greedy decoding with no constrained decoding and no LLM judge. All behavioral generations are produced under nf4 quantization, matching the training backend, so the TOST verdicts describe the deployed-quantization regime end-to-end. max\_new\_tokens $= 16$ (router) or 128 (json, extraction). The 16-token router limit truncates a few FORMAT\_BREAK outputs (6/251 at margin3). The router format gate requires the output to begin with one of the five valid labels. A 16-token truncation cannot turn an invalid output into a valid label, because all valid labels are shorter than the limit (3--6 tokens). The baseline is the step-0 SFT checkpoint on the same suite and scorer, evaluated deterministically.

\begin{table}[H]
\centering
\caption{Evaluation tasks: output format, example, and preference-data coverage.}
\label{tab:tasks}
\small
\setlength{\tabcolsep}{4pt}
\begin{tabular}{@{}>{\raggedright\arraybackslash}p{1.5cm}>{\raggedright\arraybackslash}p{4.0cm}>{\raggedright\arraybackslash}p{3.4cm}>{\raggedright\arraybackslash}p{2.5cm}@{}}
\toprule
\textbf{Task} & \textbf{Output format} & \textbf{Example} & \textbf{Pref.\ coverage}\\
\midrule
json & Structured JSON tool call (function name + typed args) & \texttt{\small \{"tool":\newline"search\_filings",\newline"arguments":\newline\{"ticker":"AAPL"\}\}} & On-target (171 tool\_calling training pairs; 200 in the full dataset)\\
router & Bare UPPER\_SNAKE\_CASE label from a fixed 5-label set (FILING\_ANALYSIS, FILING\_DIFF, COMPANY\_DEEP\_DIVE, SIMPLE\_QUERY, OUT\_OF\_SCOPE) & \texttt{\small COMPANY\_\allowbreak DEEP\_DIVE} & Off-target (zero)\\
extraction & Three-field JSON: ticker, section, filing\_date & \texttt{\small \{"ticker":"MSFT",\newline "section":"Item 7",\newline "filing\_date":\newline "2024-02-15"\}} & Off-target (zero)\\
\bottomrule
\end{tabular}
\end{table}

Scoring uses scorer v2, a frozen four-tier classification applied in this order. FORMAT\_BREAK: the output fails the task's parse gate (for router, it does not begin with a valid label, including syntactically plausible but invalid labels; for json and extraction, it is not valid JSON with the required keys). PASS: strict match to gold. SURFACE\_FORM: no strict match, but a match after canonicalization (e.g., ``Item 1'' rendered as ``1''). SEMANTIC\_ERROR: no match after canonicalization. FORMAT\_BREAK rate is also tracked separately. The primary metric is semantic accuracy (PASS $+$ SURFACE\_FORM). The extraction canonicalization rule was frozen before the confirmatory run (Appendix~\ref{app:scorer}). It matches section identifiers via regex \verb|\d+[A-Z]?|, case-folded, ignoring appended titles.

\begin{tcolorbox}[colback=gray!6,colframe=gray!55,boxrule=0.5pt,left=6pt,right=6pt,top=4pt,bottom=4pt,title={\small\textbf{The scorer artifact: strict versus canonicalized scoring}}]
\small Under strict string matching, extraction accuracy at margin3 appears to crash, but canonicalized scoring recovers it: seed0 $0.550 \to 0.981$ ($69/72 = 96\%$); 8-seed mean $0.616 \to 0.988$. The flipped items are surface-form register shifts (``Item 1'' versus ``1''), not semantic failures. This is the difference between ``catastrophic collapse at margin ${\approx}3$'' and ``capability preserved with a transient register shift.''
\end{tcolorbox}

\subsection{Suite Independence and Domain Overlap}
\label{sec:suite}
No eval prompt appears verbatim in the preference or SFT training data (exact-string match: 0/480 against the 647 preference questions and 0/480 against the 1{,}060 unique SFT user prompts; Appendix~\ref{app:hygiene}). However, both derive from EDGAR ticker$\times$section fetches, producing domain overlap by design. We report the most specific overlap available. Extraction shares a ticker$\times$section combination with a preference pair's source filing for $\ge$81/160 items (ticker$\times$section provenance recoverable for 209/647 pairs). The json task shares a ticker for $\ge$131/160 items (section is not a json field). Extraction's preservation may therefore partly reflect entity familiarity rather than format-robustness alone. Router has zero preference coverage and no ticker$\times$section overlap with the preference data, and provides the cleanest independence test; all three tasks share the SEC-filing domain. Degeneracy floors (extraction 0.006, router 0.200, json 0.0125), class balance, zero duplicate prompts, and gold self-consistency (160/160 per task) are reported in Appendix~\ref{app:hygiene}.

\subsection{Pre-Registered Analysis Plan}
The primary endpoint is per-task semantic accuracy at the final checkpoint (step 180) versus the step-0 SFT baseline, tested via TOST equivalence at $\pm5$ points. Let $\Delta = \text{acc}_{\text{final}} - \text{acc}_{\text{baseline}}$ in percentage points. TOST tests the composite null $H_0: |\Delta| \geq \delta$ against $H_1: |\Delta| < \delta$ with $\delta = 5$. Equivalently, it tests two one-sided nulls, $H_{01}: \Delta \leq -\delta$ and $H_{02}: \Delta \geq \delta$, each at $\alpha = 0.05$. Both are rejected if and only if the 90\% confidence interval for $\Delta$ lies entirely within $(-\delta, \delta)$:
\begin{equation}
\text{CI}_{90\%}(\Delta) \subset (-\delta, \delta) = (-5, 5).
\label{eq:tost}
\end{equation}
Eq.~\eqref{eq:tost} uses an open interval; the per-seed inclusion rule is $|\Delta| \leq 5$ inclusive. We chose $\pm5$ a priori. It is large enough that a true $\Delta$ of 0 passes the seed-level criterion (Eq.~\eqref{eq:seedci}) with power above 0.99 (post hoc, from the observed per-seed SDs of 2.28 on extraction and 1.56 on json), and small enough that passing is meaningful: 5 points is 8 of 160 items. The bound was frozen before any data from the confirmatory run existed. We acknowledge this is researcher judgment, not derived from an external standard.

Confidence intervals are constructed at two levels: seed-level (mean per-seed difference, $t$-based CI, df $= 7$) and item-level (bootstrap over items, seeds pooled). For a task to pass, both must satisfy Eq.~\eqref{eq:tost}, and $\ge$7 of 8 seeds must individually fall within $\pm5$. If all three tasks pass, the bounded-null headline holds; if any fails, the task-selective branch governs. No multiple-comparison correction is applied. The headline requires all tasks to pass (an intersection-union test), which is inherently conservative.

Formally, the primary metric for task $T$ is
\begin{equation}
\text{acc}_T(\theta) = \frac{1}{N} \sum_{i=1}^{N} \mathbf{1}\big[s(g_\theta(x_i)) \in \{\text{PASS}, \text{SURFACE\_FORM}\}\big],
\label{eq:acc}
\end{equation}
where $g_\theta$ is greedy generation and $s$ is scorer v2. $\text{FB}_T$ and $\text{CE}_T$ are defined identically with FORMAT\_BREAK and SEMANTIC\_ERROR, so that $\text{acc}_T + \text{FB}_T + \text{CE}_T = 1$. The behavioral change is $\Delta_T = 100\,[\text{acc}_T(\theta_{180}) - \text{acc}_T(\theta_{\text{ref}})]$ in percentage points. For the item-level interval, the per-item difference and its mean are
\begin{equation}
\Delta_i = \frac{1}{8} \sum_{s=1}^{8} c_i^{(s)} - c_i^{(\text{ref})}, \qquad \Delta = \frac{1}{N} \sum_{i=1}^{N} \Delta_i,
\label{eq:itemdiff}
\end{equation}
where $c_i^{(s)} \in \{0, 1\}$ is the correctness of item $i$ under seed $s$ and $c_i^{(\text{ref})}$ its correctness under the reference; Eq.~\eqref{eq:itemdiff} is a fraction, reported $\times 100$ in percentage points. The item-level interval bootstraps over items (10{,}000 resamples), treating the eight per-item observations as a cluster. The seed-level interval is
\begin{equation}
\bar{\Delta} \pm t_{0.95, 7} \cdot \frac{s_\Delta}{\sqrt{8}},
\label{eq:seedci}
\end{equation}
where $s_\Delta$ is the standard deviation of the eight per-seed $\Delta$ values. With $s_\Delta = 2.28$ the half-width is 1.53.

Final is saved at the fixed step 180. Margin1/3/5 are saved when the 20-sample smoothed training margin first crosses 1.0/3.0/5.0 (steps vary by seed: margin1 24--30, margin3 84--98, margin5 112--138). The smoothed margin at the save step is reported as each checkpoint's margin label. Cross-seed averages at these checkpoints pool seeds at matched margin but different optimizer steps; the primary endpoint (step 180) and step 20 diagnostic are unaffected. Step 20 is excluded from all endpoints. All outcome-to-interpretation mappings were frozen and date-stamped before the decisive runs (Notes 1--2, committed 2026-06-11 and 2026-06-18). The pre-registration was informed by pilot runs on seeds 0 and 1 and frozen before seeds 2--7 were run. Training is bitwise deterministic, so the confirmatory seeds 0 and 1 reproduce the pilot adapters: seed 0 at every checkpoint, seed 1 through margin5 (the pilot early-stopped at step 156; the confirmatory run continued to step 180). The confirmatory set therefore contains six seeds (2--7) whose outcomes were not visible at freeze time. All 8 seeds are reported; results restricted to seeds 2--7 are in Appendix~\ref{app:tost-data}. The notes were committed in \texttt{e3be4d5} (Note 1) and \texttt{62ab7c3} (Note 2, frozen), and are available in the repository. The repository was first pushed publicly on 2026-10-18; the commit dates reflect local authorship and are not independently timestamped.

\subsection{Probability-Layer Measurement}
This study uses two quantization backends. \emph{nf4} is the training and behavioral-evaluation backend (Kaggle T4): all training, checkpoint saving, behavioral generation (\S\ref{sec:graded}--\ref{sec:nonmono}), the nf4 training-pair reward decomposition (Table~\ref{tab:peraxis}, Appendix~\ref{app:reward-data}), and the nf4 held-out recomputation (Tables~\ref{tab:heldout-nf4} and~\ref{tab:heldout-nf4-perseed}). \emph{fp16} is the measurement on the unquantized base: the same adapters loaded onto the original, unquantized Qwen3.5-4B weights in fp16 (not dequantized nf4 weights), on Mac. Because the adapters were trained against the nf4 base, neither backend is privileged; nf4 reports what the optimizer experienced, fp16 reports what the adapter weights encode when the base is not quantized. Gold-logp measurement (\S\ref{sec:gold}), the GOLD\_PROTECTED paired statistic, the KL trajectory, the fp16 held-out margin (Table~\ref{tab:heldout-fp16}), and the fp16 training-pair decomposition (Tables~\ref{tab:peraxis} and~\ref{tab:fp16-train-pooled}) were performed under fp16. Absolute magnitudes are not comparable across backends, so substantive comparisons are within-backend. Cross-backend ratios are reported in the abstract, the \S4 opening, \S\ref{sec:mechanism} and \S\ref{sec:backend}; all other comparisons are within-backend.

To assess whether training selectively protects gold completions relative to incorrect alternatives, we teacher-force three completion types through each checkpoint: gold (correct), control A (well-formed, different section identifier), and control B (well-formed, different ticker). Both controls are structurally valid but semantically wrong. For each item $i$, the gold change from the SFT reference to the final checkpoint is
\begin{equation}
\Delta_{\text{gold}}^{(i)} = \ell_{\theta_{180}}(y_i^{\text{gold}} \mid x_i) - \ell_{\text{ref}}(y_i^{\text{gold}} \mid x_i),
\label{eq:dgold}
\end{equation}
with $\Delta_A^{(i)}$ and $\Delta_B^{(i)}$ defined identically for controls A and B. The paired statistic is
\begin{equation}
S = \frac{1}{N} \sum_{i=1}^{N} \left[ \Delta_{\text{gold}}^{(i)} - \tfrac{1}{2}\big(\Delta_A^{(i)} + \Delta_B^{(i)}\big) \right],
\label{eq:goldstat}
\end{equation}
with $N = 160$. A positive $S$ indicates gold deflates less than the controls on average (selective protection). The pre-registered threshold is $S \geq X = 0.05$ nats/token, with a 95\% bootstrap confidence interval excluding zero. Probability-layer confidence intervals are percentile bootstraps (10{,}000 resamples): per-seed intervals resample the 160 per-item values (for $S$, the per-item paired differences), and fp16 8-seed aggregate intervals resample the eight per-seed means. The nf4 aggregate intervals in Appendix~\ref{app:gold} are $t$-intervals across the eight per-seed means (df $= 7$). Controls A and B are extraction-specific; on json and router, gold-logp is reported as absolute $\Delta_{\text{gold}}$ without a paired baseline. A full reconciliation of all pre-registered secondaries against their status is in Appendix~\ref{app:reconcile}.

\textbf{Reported quantities.} The margin decomposes into chosen and rejected terms:
\begin{equation}
m = \beta\, r_\theta(x, y_w) - \beta\, r_\theta(x, y_l), \qquad \rho_{\text{ch}} = \frac{|\beta\, r_\theta(x, y_w)|}{|m|}.
\label{eq:decomp}
\end{equation}
The claim in \S\ref{sec:mechanism} that chosen reward contributes at most 4\% refers to $\rho_{\text{ch}}$, computed on 8-seed means at each checkpoint. For axis $a \in \{\text{style}, \text{tool\_calling}\}$, the aggregate margin and the backend ratio of mean per-pair margins $\bar{m}_a$ are
\begin{equation}
M_a = \sum_{i \in a} m_i, \qquad \kappa_a = \frac{\bar{m}_a^{(\text{nf4})}}{\bar{m}_a^{(\text{fp16})}}.
\label{eq:axis}
\end{equation}
We refer to $M_a$ as the axis's aggregate suppression pressure. It is the aggregate achieved margin, an outcome measure, not a measure of gradient magnitude, which decreases as $m$ grows ($\partial \mathcal{L}_{\text{pref}} / \partial m = -\sigma(-m)$). \S\ref{sec:backend} reports $\kappa_{\text{style}} \approx 3.3$ and $\kappa_{\text{tool}} \approx 1.2$. The smoothed trigger margin is
\begin{equation}
\tilde{m}_k = \frac{1}{20} \sum_{j=k-19}^{k} m_j,
\label{eq:smooth}
\end{equation}
over the 20 most recent training pairs (about 2.5 optimizer steps at batch size 8). Checkpoints fire when $\tilde{m}_k$ first reaches 1, 3, and 5. Token-level KL is
\begin{equation}
\text{KL}_t = \sum_{v} \pi_\theta(v \mid x, y_{<t}) \log \frac{\pi_\theta(v \mid x, y_{<t})}{\pi_{\text{ref}}(v \mid x, y_{<t})},
\label{eq:kl}
\end{equation}
where $v$ ranges over the token vocabulary, averaged over gold positions $t$ under teacher forcing.

\textbf{Reproducibility.} Code, the 647 preference pairs, the three 160-item evaluation suites, scorer v2, and all 40 checkpoint adapters are available at \url{https://github.com/yuan-phd/delta-filing}. The base model is Qwen3.5-4B (Hugging Face \texttt{Qwen/Qwen3.5-4B}). nf4 quantization uses bitsandbytes (4-bit nf4, double quantization, fp16 compute; installed unpinned via pip on Kaggle in June 2026, exact version not recorded). fp16 measurements use PyTorch MPS on Apple Silicon with \texttt{torch.float16} (no bf16 fallback).

\textbf{Use of AI tools.} OpenAI models were used as data-generation tools as described above: GPT-4o-mini generated the SFT data, the chosen and rejected style completions (chosen from EDGAR filing sections), the evaluation prompts, and the rejected-register control (\S\ref{sec:gold}), and GPT-4o adjudicated the router gold labels. The models were called through the OpenAI API by name (\texttt{gpt-4o-mini}, \texttt{gpt-4o}) without a dated snapshot, so exact model versions are not recorded; all generated data are released in the repository. Claude (Anthropic) was used as a coding and writing assistant for experiment and analysis code, LaTeX formatting, and language editing, under the author's direction. The author designed the study, reviewed all AI-assisted code and text, and takes full responsibility for the content of this paper.

\section{Results}

All 8 seeds completed 180 fixed optimizer steps without triggering the divergence guard. Under the nf4 training backend ($\beta = 0.5$), final margins range 6.08--7.68 (mean 6.73). Checkpoint margin labels are training-time smoothed values. The pooled value derived from Table~\ref{tab:peraxis} (all 549 train pairs) differs slightly (e.g.\ 6.73 smoothed versus 7.12 decomposed at final). Within each backend, held-out preference margin tracks training margin closely (nf4: 7.12 versus 6.98, within 2\%; fp16: 2.95 versus 2.78, within 6\%; Appendix~\ref{app:reward-data}). This indicates that margin growth is not driven by train-set memorization. Absolute margins differ $2.4\times$ (train) to $2.5\times$ (held-out) between backends, and up to $3.4\times$ on the style axis. \S\ref{sec:backend} analyzes this quantization sensitivity.

\subsection{Graded, Task-Selective Degradation}
\label{sec:graded}
Table~\ref{tab:tost} reports the verdicts of the primary TOST equivalence test (Eq.~\eqref{eq:tost}). The pre-registered falsifier fires: router degrades far beyond $\pm5$, the bounded-null headline is dropped, and the task-selective branch governs (that is, not all tasks are preserved, so results are reported per task).

Extraction passes all three criteria comfortably (point estimate $-0.55$, both CIs well within $\pm5$, 8/8 seeds inside the bound). A per-item flip analysis finds no systematic asymmetry: 15 gains versus 22 losses across seeds, the gains on 4 distinct items (the four items the baseline misses) and the losses on 16. Capability is preserved as the margin grows to ${\sim}7$.

The json task fails to establish equivalence, though marginally. The point estimate ($-3.52$) and seed-level CI are within $\pm5$. The equivalence test fails because the item-level pooled CI lower bound ($-5.31$) crosses $-5$. The per-seed criterion is met (7 of 8 within the band): seed 0 is outside at $-6.25$ and seed 4 sits on the inclusive boundary. We apply $|\Delta| \le 5$ inclusively; Note 2 did not specify strict versus inclusive. The per-item flips are 0 gains and 45 losses across seeds, on 17 distinct items. With only 2 of 160 items incorrect at baseline, zero gains is structurally near-guaranteed. The degradation is small in absolute magnitude.

The ceiling (baseline 98.75) caps gains, which reduces item variance; the item-level CI is wide because losses cluster on specific items. The 45 losses fall on 17 distinct items; five items account for 28 of them (one item lost on 7 of 8 seeds, two on 6 of 8). Because losses cluster within items, they are not independent across seeds; we therefore report item-level counts rather than pooled significance tests. Of the 45 losses, 35 are wrong-ticker errors in the free-text argument field (the tool name is correct), 7 are wrong-tool substitutions (6 on seed 0), and 3 are format breaks.

Router degrades catastrophically ($-30.16$, both CIs far outside $\pm5$, 0/8 seeds inside). The endpoint distribution is roughly symmetric (mean 0.636, median 0.638, range 0.488--0.769) but wide (spread 0.281). This reflects high seed-to-seed variance despite consistent damage depth (\S\ref{sec:nonmono}).

\begin{table}[H]
\centering
\caption{Primary TOST equivalence test ($\pm5$ pts, semantic accuracy at step 180 versus baseline).}
\label{tab:tost}
\small
\setlength{\tabcolsep}{3pt}
\begin{tabular}{@{}lrrrcccc@{}}
\toprule
Task & Base & Final & $\Delta$ & Seed 90\% CI & Item 90\% CI & Seeds $|\Delta|\le5$ & Verdict\\
\midrule
extraction & 97.50 & 96.95 & $-0.55$ & $[-2.07, +0.98]$ & $[-1.80, +0.86]$ & 8/8 & Equivalent\\
json & 98.75 & 95.23 & $-3.52$ & $[-4.56, -2.47]$ & $[-5.31, -1.95]$ & 7/8 & Equiv.\ not shown\\
router & 93.75 & 63.59 & $-30.16$ & $[-35.65, -24.66]$ & $[-34.92, -25.47]$ & 0/8 & Not equivalent\\
\bottomrule
\end{tabular}
\end{table}

The alignment tax is real, graded, and task-selective: undetectable (extraction, $\Delta = -0.55$), small but genuine (json, $\Delta = -3.52$), catastrophic (router, $\Delta = -30.16$). The degradation ordering is directionally consistent with each task's format-alignment to the preference data: router is worst on 8/8 seeds; extraction loses less than json on 6/8 seeds; the full ordering holds in the mean. Router's terse enum format is the most orthogonal to the dominant verbose style axis, and router degrades most. The json task has matching tool\_calling coverage and degrades modestly. Extraction, whose format is orthogonal to but not antagonistic to the style axis, is preserved. With only three tasks, we report this co-occurrence without claiming the causal link is established. Extraction also lacks direct preference coverage yet is preserved, so coverage absence alone does not predict degradation. The json task has 171 tool\_calling training pairs (200 in the full dataset) yet still degrades. The per-axis dynamics (\S\ref{sec:mechanism}) suggest a reason: tool\_calling coverage may provide partial but not complete protection against the dominant aggregate pressure from the style axis.

\subsection{Non-Monotonic Dynamics with Two Failure Modes}
\label{sec:nonmono}
Router's degradation is not monotonic in margin. Table~\ref{tab:router-traj} decomposes router accuracy at each checkpoint into its two failure modes, which show distinct aggregate trajectories. FORMAT\_BREAK rate is non-monotonic, peaking at margin ${\approx}3$ (0.196) then partly subsiding by final (0.106). Content-error rate rises (0.148 $\to$ 0.258), rising monotonically in the mean. The partial recovery in overall accuracy after margin3 is driven by format-break subsiding, not by content correctness improving.

\begin{table}[H]
\centering
\caption{Router trajectory decomposition (mean $\pm$ std over 8 seeds; nf4, training backend).}
\label{tab:router-traj}
\small
\begin{tabular}{@{}lrccc@{}}
\toprule
Checkpoint & Margin & Semantic acc & FORMAT\_BREAK rate & Content-error rate\\
\midrule
step 20 & 0.75 & $0.852 \pm 0.015$ & $0.001 \pm 0.002$ & $0.148 \pm 0.016$\\
margin1 & 1.09 & $0.816 \pm 0.030$ & $0.007 \pm 0.010$ & $0.177 \pm 0.038$\\
margin3 & 3.12 & $0.587 \pm 0.061$ & $0.196 \pm 0.089$ & $0.217 \pm 0.071$\\
margin5 & 5.13 & $0.626 \pm 0.049$ & $0.133 \pm 0.096$ & $0.241 \pm 0.083$\\
final & 6.73 & $0.636 \pm 0.077$ & $0.106 \pm 0.070$ & $0.258 \pm 0.074$\\
\bottomrule
\end{tabular}
\end{table}

Format breaks at margin ${\approx}3$ take the form of invented UPPER\_SNAKE\_CASE pseudo-labels derived from query content (e.g., TOTAL\_REVENUES\_2023\_Q2, REPORT\_PER\_ANALYSIS). The valid router labels are also UPPER\_SNAKE\_CASE, so the model generates in the correct register but invents labels outside the valid 5-category set. This is a label-vocabulary interference specific to the preference data's financial domain, not degeneration, truncation, or style bleed. This replicates across the seeds of the confirmatory run.

The margin ${\approx}3$ instability is not limited to router. On extraction, the SURFACE\_FORM register-shift rate mirrors the non-monotonic shape: it spikes at margin ${\approx}3$ (mean 0.37, range 0.03--0.77) then subsides to 0.09 by final. Under the strict scorer this register shift counts as a capability collapse. Under scorer v2's canonicalization the apparent errors are surface-form shifts, not semantic failures (see the scorer artifact in \S\ref{sec:tasks}). This is the paper's most concrete demonstration of why artifact-aware scoring is necessary.

\begin{figure}[t]
\centering
\includegraphics[width=0.74\linewidth]{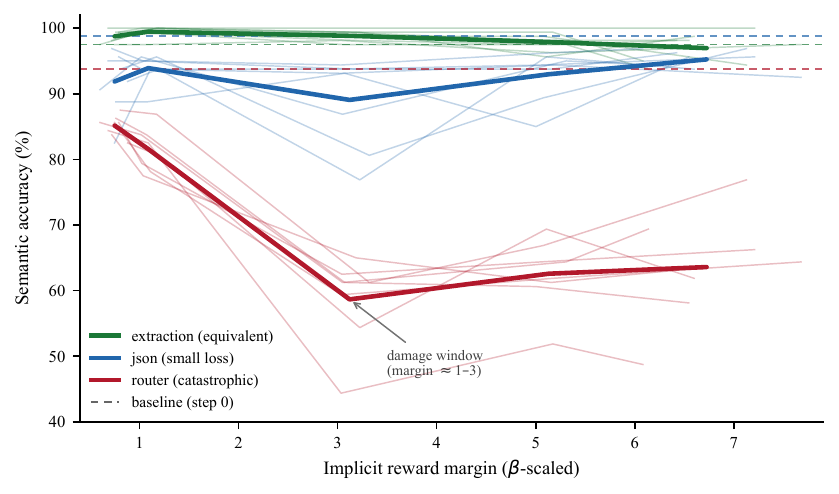}
\caption{Per-seed margin-vs-capability trajectories for all three tasks (nf4, training backend). Thin lines are individual seeds, each drawn at that seed's own smoothed margin at each checkpoint; thick lines are the 8-seed mean at the mean checkpoint margins; dashed lines are per-task baselines. Extraction (green) stays near baseline; json (blue) is volatile mid-trajectory but converges narrowly; router (red) fans out, with the sampled minimum at margin3 on 6/8 seeds.}
\label{fig:traj}
\end{figure}

Router accuracy reaches its trajectory minimum at margin3 on 6 of 8 seeds (mean 0.587, range 0.444--0.650). Two seeds drift slightly lower afterward. Degradation begins before the margin 1--3 interval: step 20 already shows $-8.6$ pts from baseline (28\% of the net $-30.2$ pt loss), and margin1 accounts for $-12.2$ pts (40\%). Training uses a constant learning rate with no warmup, so some of this early loss may reflect optimization transient rather than margin-driven damage; we did not run a warmup ablation. The largest observed drop occurs between the margin1 and margin3 checkpoints. The sampled minimum is at margin3 on 6/8 seeds; at this checkpoint resolution, the continuous trough lies somewhere between margin1 and margin5. Despite consistent trough depth, recovery is highly seed-dependent. Endpoint accuracy ranges 0.488 (seed5, near-zero recovery) to 0.769 (seed1), and the final format-break rate ranges 0.019--0.256. The damage is reproducible; the recovery is stochastic. Figure~\ref{fig:traj} shows per-seed margin-vs-capability trajectories. Extraction forms a tight band near baseline, json shows mid-trajectory volatility but a narrow endpoint range, and router forms a wide fan. The per-seed spread is itself the result: damage depth is consistent, recovery is not.

\subsection{Mechanism: Rejected Collapse and Axis-Dependent Backend Sensitivity}
\label{sec:mechanism}
\begin{figure}[t]
\centering
\includegraphics[width=0.7\linewidth]{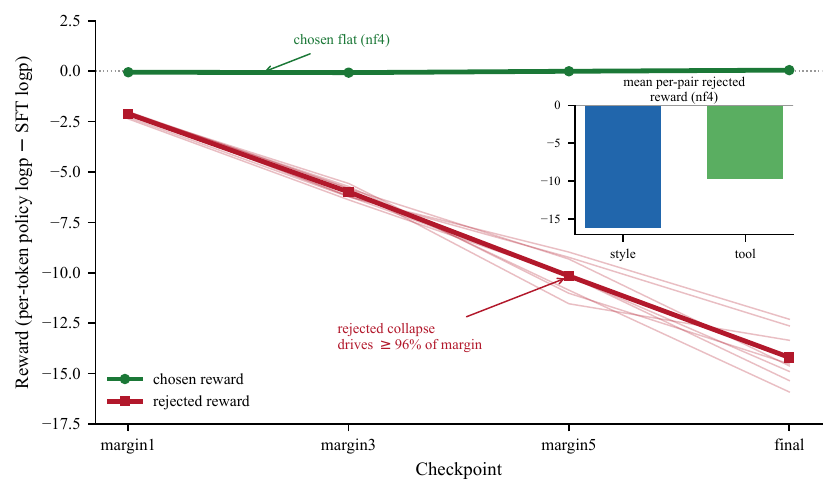}
\caption{Reward decomposition into chosen and rejected components across checkpoints (nf4, training backend; thin lines per seed, thick lines 8-seed mean). Chosen reward stays flat near zero, consistent with the retention term anchoring it; rejected reward collapses, accounting for almost all (${\geq}96\%$) of the margin growth. Inset: mean per-pair rejected reward (nf4) at the final checkpoint (style versus tool\_calling).}
\label{fig:reward}
\end{figure}

We decompose the margin $m$ (Eq.~\eqref{eq:pref}) into its chosen and rejected reward components. Chosen reward is approximately stationary at the final checkpoint under both backends (nf4: $+0.052$; fp16: $-0.028$). This is consistent with the SFT retention term (weight 0.5) anchoring the likelihood of the chosen completion. Under fp16, chosen reward dips to $-0.173$ at margin3 before recovering. This is a transient decline in Rafailov et al.'s predicted direction, which the retention term dampens but does not fully block. nf4 shows a $2.4\times$ shallower dip at the same checkpoint ($-0.072$ versus $-0.173$). The chosen reward's contribution ($\rho_{\text{ch}}$, Eq.~\eqref{eq:decomp}) is $\le4\%$ in magnitude at every checkpoint, so margin growth is accounted for almost entirely (${\geq}96\%$) by rejected-completion collapse. Even with chosen reward explicitly anchored, the behavioral alignment tax remains: router loses 30 points and json loses 3.52. The margin decomposition shows where the margin comes from, not where the damage comes from. The DPO gradient on chosen, the DPO gradient on rejected, and the SFT gradient on chosen all move the weights, and attributing the damage to one channel requires a no-retention control (\S\ref{sec:limits}). Implications for Cal-DPO and Rafailov et al.\ are discussed in \S\ref{sec:priorwork}. Figure~\ref{fig:reward} (nf4, training backend) shows the per-seed trajectories.

The per-axis decomposition reveals that per-pair ordering is not backend-invariant. Table~\ref{tab:peraxis} reports per-axis means at the final checkpoint under both backends.

\begin{table}[H]
\centering
\caption{Per-axis reward decomposition at final checkpoint (8-seed mean; 549 training pairs: 378 style, 171 tool\_calling).}
\label{tab:peraxis}
\small
\setlength{\tabcolsep}{4.5pt}
\begin{tabular}{@{}lrrrrrrr@{}}
\toprule
Axis & Pairs & Margin (nf4) & Margin (fp16) & $r_{\text{ch}}$ (nf4) & $r_{\text{rej}}$ (nf4) & $r_{\text{ch}}$ (fp16) & $r_{\text{rej}}$ (fp16)\\
\midrule
style & 378 & $8.119$ & $2.466$ & $+0.012$ & $-16.226$ & $-0.024$ & $-4.955$\\
tool\_calling & 171 & $4.923$ & $4.021$ & $+0.142$ & $-9.704$ & $-0.038$ & $-8.080$\\
\bottomrule
\end{tabular}
\end{table}

Under nf4, the training backend, style dominates on both per-pair margin (8.12 versus 4.92) and pair count. The aggregate margin is 3069 versus 842, a $3.65\times$ ratio. These are the margins the optimizer actually experienced. Under fp16 measurement on the unquantized base, the per-pair ordering reverses: tool\_calling is higher (4.02 versus 2.47). Style still produces more aggregate margin through pair count (932 versus 688, $1.36\times$). The divergence is concentrated in the style axis's rejected completions (nf4 $-16.2$ versus fp16 $-5.0$; tool\_calling $-9.7$ versus $-8.1$).

Under both backends, style produces more aggregate margin. The degradation ordering is consistent with this aggregate pressure pattern: the task most format-orthogonal to the verbose style axis (router) absorbs the most damage.

This per-pair distinction is tested on held-out data. On 98 held-out pairs (69 style, 29 tool\_calling), tool\_calling margin exceeds style margin at every checkpoint under fp16 (8/8 seeds; Appendix~\ref{app:reward-data}, Table~\ref{tab:heldout-fp16}). Under nf4, tool\_calling leads at margin1--3 (8/8 seeds; Table~\ref{tab:heldout-nf4-perseed}) but style overtakes by margin5, consistent with the axis-dependent inflation measured on training pairs at final. The pre-registered prediction that style margin would exceed tool\_calling margin at every checkpoint (Note 2, secondary \#3) is falsified under both backends: the prediction holds at 0/4 checkpoints under fp16 and 2/4 under nf4. The final-checkpoint per-pair ordering is backend-dependent.

\subsection{Gold-Logp Dynamics Are Task-Selective}
\label{sec:gold}
We next examine whether gold completions on evaluation items are stable and whether the pattern generalizes across tasks. All $\Delta_{\text{gold}}$ values are computed in fp16 on Mac. Adapters were trained under nf4 on T4, but adapter weights are stored in full precision, so loading nf4-trained adapters onto an fp16 base is valid. Within-backend comparisons are internally consistent; absolute magnitudes may differ from nf4.

\begin{table}[H]
\centering
\caption{Same-backend fp16 $\Delta_{\text{gold}}$ (gold\_logp(final) $-$ gold\_logp(SFT); nats/token), 8-seed mean. 95\% CI: percentile bootstrap over the eight per-seed means.}
\label{tab:gold}
\small
\begin{tabular}{@{}lccc@{}}
\toprule
Task & $\Delta_{\text{gold}}$ & 95\% CI & Behavioral $\Delta$\\
\midrule
extraction & $-0.15$ & $[-0.18, -0.13]$ & $-0.55$ pts (equivalent)\\
json & $-1.04$ & $[-1.20, -0.89]$ & $-3.52$ pts (equiv.\ not shown)\\
router & $-1.47$ & $[-1.62, -1.35]$ & $-30$ pts (not equivalent)\\
\bottomrule
\end{tabular}
\end{table}

Gold deflation is task-selective (Table~\ref{tab:gold}): extraction deflates modestly ($-0.15$ nats/token), json and router substantially ($-1.04$, $-1.47$). The ordering (extraction least, json intermediate, router most) is directionally consistent with the degradation ordering. It holds in the mean with non-overlapping confidence intervals; seed 0 reverses json and router ($-1.505$ versus $-1.259$; Appendix~\ref{app:gold}). The relationship is highly non-linear, however: router's gold deflates only $1.4\times$ more than json's, but its behavioral degradation is far larger ($-30$ versus $-3.52$ points). Gold-logp magnitude alone does not explain the size of the behavioral losses. The json task shows substantial deflation ($-1.04$) but only minor behavioral loss ($-3.52$ pts). This suggests that structured output formats with redundant constraints (typed fields, function signatures) may maintain correct greedy-decode behavior despite large probability-mass redistribution. The json loss pattern is consistent with this: 35 of 45 losses land on the one free-text argument (ticker), not on the typed tool-selection field. Router's terse single-label enum format lacks this structural buffer.

The task-selective ordering is the same under both backends, though the nf4 comparison also changes the baseline checkpoint. Using the nf4 records of the confirmatory run (step 20$\to$final), the ordering extraction $\ll$ json $<$ router is identical (nf4 $\Delta_{\text{gold}}$: extraction $-0.008$, json $-0.22$, router $-1.59$; non-overlapping CIs; Appendix~\ref{app:gold}). Absolute magnitudes diverge: extraction's deflation nearly vanishes under nf4 (from $-0.15$ to $-0.008$), json compresses ${\sim}5\times$, and router is essentially unchanged. Whether it reflects quantization-dependent sensitivity, baseline differences, or an interaction is unresolved. The task-selective ordering, which is the load-bearing claim, holds under both backends.

\begin{figure}[t]
\centering
\includegraphics[width=0.62\linewidth]{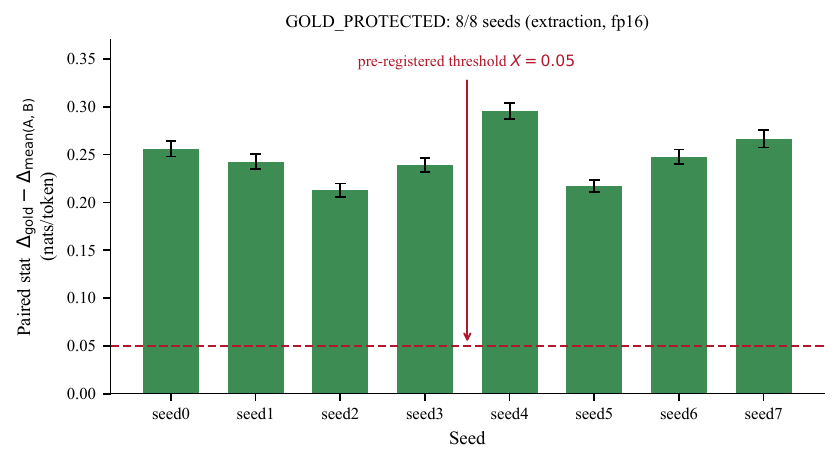}
\caption{Paired GOLD\_PROTECTED statistic per seed with 95\% CI (extraction, fp16). All 8 bars clear the pre-registered threshold $X = 0.05$ (dashed).}
\label{fig:gold}
\end{figure}

On extraction we can go further. We teacher-force gold alongside control A (different section identifier) and control B (different ticker) through each seed's final checkpoint. This yields the paired GOLD\_PROTECTED statistic of Eq.~\eqref{eq:goldstat} (Table~\ref{tab:protected}, Figure~\ref{fig:gold}). All 8 seeds return GOLD\_PROTECTED: every stat exceeds the pre-registered threshold ($X = 0.05$) with CIs far from zero. Gold is selectively protected. It declines modestly ($-0.10$ to $-0.23$ nats/token) while incorrect controls collapse $2.1\times$ to $3.5\times$ more ($-0.34$ to $-0.47$). Gold's average advantage over the controls increases. The paired selective-protection statistic was measured only under fp16. Because the controls are extraction-specific, whether gold is relatively protected on json and router is untested (Table~\ref{tab:gold} reports their absolute $\Delta_{\text{gold}}$ only). Two seeds from a preliminary probe run (seeds 0 and 1, nf4) show no evidence of the catastrophic displacement mode that Razin et al.\ flag with high CHES scores. On both seeds, a third control (rejected-register text) falls sharply ($-4.6$ to $-5.1$ nats/token). Control C was not in the confirmatory run or the Mac runs. Finally, KL(policy $\|$ reference) on extraction gold completions rises from 0.022 at margin1 to 0.082 at final (Appendix~\ref{app:gold}). This confirms mild policy drift that plateaus as margin grows from ${\sim}5$ to ${\sim}7$. KL is computed on extraction gold only, so drift on router may be larger.

\begin{table}[H]
\centering
\caption{Paired GOLD\_PROTECTED statistic (fp16, extraction, 160 items per seed).}
\label{tab:protected}
\small
\begin{tabular}{@{}lcccc@{}}
\toprule
Seed & Stat & 95\% CI & $\Delta_{\text{gold}}$ & $\Delta_{\text{ctrl}}$\\
\midrule
0 & $+0.256$ & $[+0.248, +0.264]$ & $-0.102$ & $-0.358$\\
1 & $+0.242$ & $[+0.235, +0.250]$ & $-0.226$ & $-0.469$\\
2 & $+0.213$ & $[+0.205, +0.220]$ & $-0.147$ & $-0.360$\\
3 & $+0.239$ & $[+0.232, +0.246]$ & $-0.102$ & $-0.341$\\
4 & $+0.295$ & $[+0.287, +0.304]$ & $-0.150$ & $-0.446$\\
5 & $+0.217$ & $[+0.211, +0.224]$ & $-0.194$ & $-0.411$\\
6 & $+0.248$ & $[+0.240, +0.255]$ & $-0.138$ & $-0.386$\\
7 & $+0.267$ & $[+0.257, +0.276]$ & $-0.168$ & $-0.435$\\
\bottomrule
\end{tabular}
\end{table}

\subsection{Backend Sensitivity of QLoRA Margins}
\label{sec:backend}
Implicit reward margins under QLoRA differ up to $3.4\times$ between nf4 measurement and fp16 measurement on the unquantized base, for the same adapters. Both backends use the same SFT adapter as reference, scored within each backend. The divergence is axis-dependent. On training pairs, $\kappa_{\text{style}} \approx 3.3$ and $\kappa_{\text{tool}} \approx 1.2$ (Eq.~\eqref{eq:axis}); on held-out pairs, style inflates $3.4\times$ and tool\_calling $1.2\times$. Style-rejected completions are long degraded-text sequences, while tool\_calling-rejected completions are short tool-swaps. Length and content co-vary with axis, and the cause of the differential inflation is not isolated. Final-checkpoint anchoring of chosen reward is backend-robust, but the mid-training trajectory is not: the margin3 transient is $2.4\times$ deeper under fp16 ($-0.173$ versus nf4's $-0.072$).

In our measurements, this is directionally opposite to the gold-logp nf4/fp16 discrepancy (\S\ref{sec:gold}), where nf4 magnitudes are smaller for extraction ($-0.008$ versus $-0.15$) and json (${\sim}5\times$) and comparable for router. However, the gold-logp comparison is baseline-confounded (SFT versus step 20), while the margin comparison uses the same SFT adapter scored within each backend. The contrast is therefore suggestive, not established. Chosen-flatness at the final checkpoint and aggregate ordering (style $>$ tool under both backends) are backend-invariant; per-pair ordering is not. Margin-based monitoring of QLoRA-trained models is therefore backend-sensitive. Per-pair and per-axis comparisons require same-backend measurement.

\section{Discussion}

\subsection{A Two-Level Mechanistic Account}
The behavioral and probability-layer findings form a coherent account, though not a causally proven one. At the training level, chosen-completion likelihood stays anchored, consistent with the retention term, and almost the entire margin (${\geq}96\%$) is accounted for by rejected suppression (\S\ref{sec:mechanism}). The style axis contributes more aggregate suppression than the correctness axis under both backends ($3.65\times$ under nf4, $1.36\times$ under fp16). This is consistent with the observation that the task most orthogonal in format to the verbose style axis (router) absorbs the most damage, while the task with direct tool\_calling coverage (json) absorbs less. Per-pair ordering, by contrast, is backend-dependent. At the probability level, gold-logp deflation is task-selective and directionally consistent with the degradation ordering. Extraction is the best-preserved task, and its gold completions share entity vocabulary with training-pair completions (\S\ref{sec:suite}; measures differ by task). On extraction, gold is selectively protected relative to incorrect controls (8/8 seeds). This is consistent with Ren and Sutherland's squeezing effect concentrating mass onto gold tokens that share entity vocabulary with training pairs. Router's dissimilar terse format may not benefit from this effect. Both backends' chosen-reward minima sit at margin3, coinciding with the behavioral instability window (FORMAT\_BREAK 0.196, SURFACE\_FORM 0.37). Behavioral metrics are nf4-generated, and the fp16 chosen dip is a measurement on the unquantized base, not the source of the behavioral instability. No causal claim is made.

The two levels do not form a single causal chain. Gold-logp magnitude is non-linearly related to behavior: router's gold deflates $1.4\times$ more than json's but degrades far more ($-30$ versus $-3.52$ points). The degradation ordering is better explained by format-alignment to the aggregate pressure from the style axis than by gold-logp dynamics. We present the two levels as complementary, without claiming one causes the other. Rejected collapse is consistent with which tasks are affected; gold-logp dynamics describe how probability mass redistributes.

\subsection{Falsification Chronology}
\label{sec:chronology}
The pre-registered plan forced three falsifications and two refinements. Each produced a more precise characterization than the hypothesis it replaced. Three further hypotheses were abandoned during design (Appendix~\ref{app:reconcile}): a confounded causal-clamp experiment, router as the sole primary endpoint after a non-monotonic transient appeared (its trajectory analysis became a secondary), and a targeted-displacement probe that pointed to global drift.

Three hypotheses were falsified outright, two in the confirmatory run and one in the pilot. First, the bounded-null headline (all three tasks preserved within $\pm5$) was rejected when the falsifier fired on json and router (\S\ref{sec:graded}). Second, an apparent extraction collapse at margin ${\approx}3$, observed in the pilot, was traced to a scorer artifact before the confirmatory run (see the scorer artifact in \S\ref{sec:tasks}): the apparent errors were surface-form register shifts. The canonicalization rule was frozen in response, and the strict-scorer measurement was discarded, not refined. Third, the pre-registered prediction that style-axis held-out margin would exceed tool\_calling margin at every checkpoint is falsified under both backends (fp16: 0/4 checkpoints; nf4: 2/4). The final-checkpoint per-pair ordering is backend-dependent (style $>$ tool under nf4, tool $>$ style under fp16). The falsification reveals that per-pair ordering is not backend-invariant; aggregate style dominance holds under both.

Two characterizations were refined. The pilot's margin-independent early drop on router became a damage window at margin ${\approx}1$--3 with two failure modes with distinct aggregate trajectories (transient format-break $+$ monotonic content-error in the mean), followed by stabilization (\S\ref{sec:nonmono}). The initial gold barely moves characterization came from nf4 ($|\Delta_{\text{gold}}| \le 0.01$). It was refined to ``gold is selectively protected but task-selectively deflates'' once fp16 same-backend measurement revealed substantial json and router deflation (\S\ref{sec:gold}). Each pre-registered gate forced a more precise account. The result is more nuanced than any single prior framing predicted.

\subsection{Engagement with Prior Work}
\label{sec:priorwork}
Our data engages three displacement-literature claims. One caveat applies: the objective we study includes an SFT retention term (the RPO formulation; Liu et al.\ \citep{liu2024rpo}; weight 0.5) on chosen completions, making it DPO $+$ retention, not vanilla DPO. These are observations in one regime, not tests of the cited methods or of their frameworks in general. \emph{Rafailov et al.}\ predict from their Q-function framework that chosen log-prob should decrease under DPO from an SFT-initialized model. The retention term directly counteracts this. The chosen-flatness we observe ($-0.072$ to $+0.052$ over 7 units of margin growth) is therefore at least partly a consequence of the training objective, not purely an empirical finding about DPO dynamics. It is consistent with the retention term anchoring chosen reward in this regime. It cannot speak to whether chosen reward would decline under standard DPO without retention. The fp16 measurement on the unquantized base reveals a transient mid-training chosen dip ($-0.173$ at margin3, recovering to $-0.028$ by final). This is a residual decline in the direction Rafailov et al.\ predict, dampened but not fully blocked by the retention term. Under nf4 the dip is $2.4\times$ shallower ($-0.072$). The retention term largely counteracts the predicted decline; measurement on the unquantized base reveals the residual. \emph{Cal-DPO} adds a calibration term to prevent chosen-reward decline; the retention term achieves a similar anchoring by a different mechanism. That chosen reward is anchored at the endpoint does not demonstrate Cal-DPO's concern is unfounded. It is consistent with the retention term largely addressing the concern. Even with chosen reward anchored, the behavioral tax remains, with almost the entire margin (${\geq}96\%$) accounted for by rejected suppression; which gradient channel produces the behavioral damage is untested. \emph{IPO} was designed to bound the margin, treating unbounded growth as the core pathology. Our margin grows to ${\sim}7$ with graded, format-selective damage: the damage is task-selective rather than uniform. We did not train with IPO; in this regime, damage is concentrated in the margin 1--3 window and is format-selective.

\subsection{Format-Alignment: What It Explains and What It Does Not}
The degradation ordering (extraction preserved $\to$ json small loss $\to$ router catastrophic) is directionally consistent with format-alignment to the preference data's style axis: router is worst on 8/8 seeds; extraction loses less than json on 6/8 seeds; the full ordering holds in the mean. With only three tasks, however, we report this co-occurrence without claiming a causal mechanism. Two limits deserve attention. First, extraction has zero direct preference coverage yet is preserved, so coverage absence alone does not predict degradation. The distinguishing factor may be format-orthogonality, domain overlap (\S\ref{sec:suite}), or the retention term anchoring chosen completions that share extraction's entity vocabulary. All three co-vary and cannot be disentangled with three tasks. Second, json has 171 on-target tool\_calling training pairs (200 in the full dataset) yet degrades slightly. Tool\_calling coverage may protect against the correctness axis but not against the dominant aggregate pressure from the style axis. If so, the damage and the protection operate on different axes. The json task shows substantial gold deflation ($-1.04$) but only minor behavioral loss ($-3.52$ points). This suggests redundant-constraint formats may buffer greedy-decode behavior against probability-mass redistribution (see \S\ref{sec:graded} for the loss breakdown). But extraction is also structured JSON yet deflates far less ($-0.15$), so format redundancy alone does not explain the difference; format-alignment co-varies. With two structured tasks and two co-varying variables, this hypothesis is illustrative, not testable.

\subsection{Practical Implications}
Three implications follow for practitioners running length-normalized DPO on small models with structured-output tasks. (1)~Monitor format-break rate at margin ${\approx}3$ (on this setup's nf4 smoothed-margin scale), not just endpoint accuracy. The transient spike is where damage concentrates, and it may be invisible at the final checkpoint. (2)~Audit preference-data composition for style-axis aggregate dominance. The degradation ordering is consistent with aggregate training pressure under both backends, though per-pair ordering is backend-dependent and per-axis comparisons require same-backend measurement. (3)~Consider including an SFT retention term. Its anchoring is endpoint-level: mid-training chosen reward can dip transiently under measurement on the unquantized base. The no-retention counterfactual is untested.

\subsection{Limitations}
\label{sec:limits}
This study operates within explicit scope conditions: sparse LoRA on 8/32 layers (0.034\% of parameters; not full fine-tuning), a single 4B model, length-normalized rewards with an SFT retention term (weight 0.5), easily-separable synthetic preferences, and three structured exact-format tasks. We do not study safety or refusal tasks, where Razin et al.\ demonstrated catastrophic displacement. Under naturalistic human preferences with higher chosen-rejected similarity (high CHES scores), the displacement dynamics Razin et al.\ characterize may produce different outcomes. Results may not generalize beyond these conditions. The no-retention counterfactual ($\lambda = 0$ in Eq.~\eqref{eq:total}) is untested; whether chosen reward declines and whether behavioral damage changes without the retention term are open questions.

Three pre-registered secondaries (\#3, \#4, \#8) were computed post-hoc from saved adapters rather than logged during training: \#4 under nf4 (Kaggle T4) with an fp16 measurement on the unquantized base (Mac); \#3 under fp16 (Mac) and nf4 (Kaggle T4); \#8 under fp16 (Mac). Separately, the paired GOLD\_PROTECTED statistic is extraction-only by design (controls A/B are task-specific). The format-alignment mechanism rests on $n = 3$ tasks. This is insufficient to distinguish format-orthogonality from coverage, domain overlap, or task difficulty. The checkpoint schedule is coarse: the largest observed drop occurs between the margin1 and margin3 checkpoints, but the sampled minimum (margin3 on 6/8 seeds) places the continuous trough only somewhere between margin1 and margin5. Gold-logp absolute magnitudes differ substantially between fp16 and nf4. The task-selective ordering is the same under both, but absolute values are not directly comparable (\S\ref{sec:gold}, \S\ref{sec:backend}, Appendix~\ref{app:gold}).

\section{Conclusion}
The alignment tax under length-normalized DPO with an SFT retention term is real but graded and task-selective: extraction is equivalent within $\pm5$ pts under nf4 evaluation, json loses little, and router degrades severely. Router's loss concentrates in a margin 1--3 damage window, where a transient format-break rate and a content-error rate that rises monotonically in the mean follow distinct aggregate trajectories. Chosen reward stays anchored, consistent with the retention term, and margin growth is accounted for almost entirely (${\geq}96\%$) by suppression of rejected completions; a format-alignment hypothesis is consistent with, but not established by, the three-task ordering. Gold-logp deflation is task-selective; on extraction, gold's average advantage over incorrect controls increases under fp16, 8/8 seeds. Implicit reward margins differ up to $3.4\times$ between nf4 and unquantized fp16 measurement, with the gap depending on the preference axis. Artifact-aware scoring and pre-registered TOST equivalence testing prevented a false-collapse conclusion, and a disclosed chronology of three falsified hypotheses and two refined characterizations documents how pre-registration refined initially simpler claims. These findings hold for sparse LoRA on 8/32 layers, a single 4B model, length-normalized rewards with an SFT retention term, and easily-separable synthetic preferences. They may not transfer to full fine-tuning, naturalistic preferences, or safety-critical tasks.

\section*{Acknowledgements}
I thank Dr. Jing Su for helpful comments on two rounds of the manuscript. No external funding; no competing interests.

\bibliographystyle{plainnat}
\bibliography{references}

\begin{thebibliography}{16}
\providecommand{\natexlab}[1]{#1}
\providecommand{\url}[1]{\texttt{#1}}
\expandafter\ifx\csname urlstyle\endcsname\relax
  \providecommand{\doi}[1]{doi: #1}\else
  \providecommand{\doi}{doi: \begingroup \urlstyle{rm}\Url}\fi

\bibitem[Azar et~al.(2024)Azar, Guo, Piot, Munos, Rowland, Valko, and Calandriello]{azar2024ipo}
Mohammad~Gheshlaghi Azar, Zhaohan~Daniel Guo, Bilal Piot, Remi Munos, Mark Rowland, Michal Valko, and Daniele Calandriello.
\newblock A general theoretical paradigm to understand learning from human preferences.
\newblock In \emph{International Conference on Artificial Intelligence and Statistics (AISTATS)}, 2024.

\bibitem[Bao and Abisado(2026)]{bao2026pareto}
Shenghui Bao and Mideth Abisado.
\newblock Pareto-aware dual-preference optimization for task-oriented dialogue.
\newblock \emph{Symmetry}, 18\penalty0 (2):\penalty0 372, 2026.

\bibitem[Baxi(2025)]{baxi2025scoring}
Rahul Baxi.
\newblock Separating constraint compliance from semantic accuracy: A novel benchmark for evaluating instruction-following under compression.
\newblock \emph{arXiv preprint arXiv:2512.17920}, 2025.

\bibitem[Lewkowycz et~al.(2022)Lewkowycz, Andreassen, Dohan, Dyer, Michalewski, Ramasesh, Slone, Anil, Schlag, Gutman-Solo, Wu, Neyshabur, Gur-Ari, and Misra]{lewkowycz2022minerva}
Aitor Lewkowycz, Anders Andreassen, David Dohan, Ethan Dyer, Henryk Michalewski, Vinay Ramasesh, Ambrose Slone, Cem Anil, Imanol Schlag, Theo Gutman-Solo, Yuhuai Wu, Behnam Neyshabur, Guy Gur-Ari, and Vedant Misra.
\newblock Solving quantitative reasoning problems with language models.
\newblock In \emph{Advances in Neural Information Processing Systems (NeurIPS)}, 2022.

\bibitem[Li(2026)]{li2026oxrl}
Xiaoyi Li.
\newblock Do post-training algorithms actually differ? a controlled study across model scales uncovers scale-dependent ranking inversions.
\newblock \emph{arXiv preprint arXiv:2603.19335}, 2026.

\bibitem[Liu et~al.(2024)Liu, Lu, Zhang, Liu, Guo, Yang, Blanchet, and Wang]{liu2024rpo}
Zhihan Liu, Miao Lu, Shenao Zhang, Boyi Liu, Hongyi Guo, Yingxiang Yang, Jose Blanchet, and Zhaoran Wang.
\newblock Provably mitigating overoptimization in {RLHF}: Your {SFT} loss is implicitly an adversarial regularizer.
\newblock In \emph{Advances in Neural Information Processing Systems (NeurIPS)}, 2024.
\newblock arXiv:2405.16436.

\bibitem[Manzoni(2026)]{manzoni2026tost}
Andrea Manzoni.
\newblock {MoE}-sieve: Routing-guided {LoRA} for efficient {MoE} fine-tuning.
\newblock \emph{arXiv preprint arXiv:2603.24044}, 2026.

\bibitem[Meng et~al.(2024)Meng, Xia, and Chen]{meng2024simpo}
Yu~Meng, Mengzhou Xia, and Danqi Chen.
\newblock {SimPO}: Simple preference optimization with a reference-free reward.
\newblock In \emph{Advances in Neural Information Processing Systems (NeurIPS)}, 2024.
\newblock arXiv:2405.14734.

\bibitem[Pang et~al.(2024)Pang, Yuan, Cho, He, Sukhbaatar, and Weston]{pang2024iterative}
Richard~Yuanzhe Pang, Weizhe Yuan, Kyunghyun Cho, He~He, Sainbayar Sukhbaatar, and Jason Weston.
\newblock Iterative reasoning preference optimization.
\newblock \emph{arXiv preprint arXiv:2404.19733}, 2024.

\bibitem[Pipano et~al.(2026)Pipano, Sabach, Asadi, and Ghavamzadeh]{pipano2026displacement}
Idan Pipano, Shoham Sabach, Kavosh Asadi, and Mohammad Ghavamzadeh.
\newblock Displacement-resistant extensions of {DPO} with nonconvex $f$-divergences.
\newblock In \emph{International Conference on Learning Representations (ICLR)}, 2026.
\newblock arXiv:2602.06788.

\bibitem[Pollanen(2026)]{pollanen2026margin}
Marco Pollanen.
\newblock The viscosity of logic: Phase transitions and hysteresis in {DPO} alignment.
\newblock \emph{arXiv preprint arXiv:2601.17260}, 2026.

\bibitem[Rafailov et~al.(2024)Rafailov, Hejna, Park, and Finn]{rafailov2024qfunction}
Rafael Rafailov, Joey Hejna, Ryan Park, and Chelsea Finn.
\newblock From $r$ to $q^*$: Your language model is secretly a {Q}-function.
\newblock In \emph{Conference on Language Modeling (COLM)}, 2024.

\bibitem[Razin et~al.(2025)Razin, Malladi, Bhaskar, Chen, Arora, and Hanin]{razin2025displacement}
Noam Razin, Sadhika Malladi, Adithya Bhaskar, Danqi Chen, Sanjeev Arora, and Boris Hanin.
\newblock Unintentional unalignment: Likelihood displacement in direct preference optimization.
\newblock In \emph{International Conference on Learning Representations (ICLR)}, 2025.

\bibitem[Ren and Sutherland(2025)]{ren2025squeezing}
Yi~Ren and Danica~J. Sutherland.
\newblock Learning dynamics of {LLM} finetuning.
\newblock In \emph{International Conference on Learning Representations (ICLR)}, 2025.

\bibitem[Xiao et~al.(2024)Xiao, Yuan, Zhu, Li, and Honavar]{xiao2024caldpo}
Teng Xiao, Yige Yuan, Huaisheng Zhu, Mingxiao Li, and Vasant~G. Honavar.
\newblock Cal-{DPO}: Calibrated direct preference optimization for language model alignment.
\newblock In \emph{Advances in Neural Information Processing Systems (NeurIPS)}, 2024.

\bibitem[Young(2026)]{young2026geometry}
Robin Young.
\newblock What is the alignment tax?
\newblock \emph{arXiv preprint arXiv:2603.00047}, 2026.

\end{thebibliography}

\appendix
\begin{center}
\vspace{0.5\baselineskip}
{\large\bfseries Appendices}
\end{center}

\section{Full Per-Seed TOST Data}
\label{app:tost-data}
Complete data behind Table~\ref{tab:tost}. Per-task semantic accuracy (\%) at each checkpoint, all 8 seeds, plus the step-0 SFT baseline. $\Delta = \text{final} - \text{baseline}$ (percentage points). The final column is the per-seed point check $|\Delta| \le 5$ (PASS/FAIL), not a per-seed TOST.

\begin{table}[H]
\centering
\caption{extraction: per-seed semantic accuracy (\%).}
\small
\begin{tabular}{@{}lrrrrrrrc@{}}
\toprule
Seed & Baseline & step 20 & margin1 & margin3 & margin5 & final & $\Delta$ & $|\Delta| \le 5$\\
\midrule
0 & 97.50 & 97.50 & 97.50 & 98.12 & 96.25 & 97.50 & $+0.00$ & PASS\\
1 & 97.50 & 98.75 & 100.00 & 98.75 & 98.75 & 94.38 & $-3.12$ & PASS\\
2 & 97.50 & 99.38 & 99.38 & 99.38 & 98.12 & 98.12 & $+0.62$ & PASS\\
3 & 97.50 & 98.12 & 99.38 & 98.12 & 97.50 & 98.12 & $+0.62$ & PASS\\
4 & 97.50 & 99.38 & 100.00 & 99.38 & 95.62 & 98.75 & $+1.25$ & PASS\\
5 & 97.50 & 100.00 & 100.00 & 99.38 & 99.38 & 95.00 & $-2.50$ & PASS\\
6 & 97.50 & 97.50 & 99.38 & 97.50 & 97.50 & 93.75 & $-3.75$ & PASS\\
7 & 97.50 & 99.38 & 100.00 & 100.00 & 100.00 & 100.00 & $+2.50$ & PASS\\
\bottomrule
\end{tabular}
\end{table}

\begin{table}[H]
\centering
\caption{json: per-seed semantic accuracy (\%).}
\small
\begin{tabular}{@{}lrrrrrrrc@{}}
\toprule
Seed & Baseline & step 20 & margin1 & margin3 & margin5 & final & $\Delta$ & $|\Delta| \le 5$\\
\midrule
0 & 98.75 & 82.50 & 93.75 & 93.12 & 94.38 & 92.50 & $-6.25$ & FAIL\\
1 & 98.75 & 93.75 & 95.62 & 80.62 & 89.38 & 96.88 & $-1.88$ & PASS\\
2 & 98.75 & 88.75 & 88.75 & 93.12 & 85.00 & 95.62 & $-3.12$ & PASS\\
3 & 98.75 & 96.88 & 95.00 & 93.75 & 93.75 & 96.25 & $-2.50$ & PASS\\
4 & 98.75 & 91.88 & 93.75 & 76.88 & 95.62 & 93.75 & $-5.00$ & PASS\\
5 & 98.75 & 95.00 & 95.00 & 94.38 & 96.25 & 96.88 & $-1.88$ & PASS\\
6 & 98.75 & 90.62 & 95.62 & 86.88 & 95.00 & 94.38 & $-4.38$ & PASS\\
7 & 98.75 & 95.62 & 93.75 & 93.75 & 94.38 & 95.62 & $-3.12$ & PASS\\
\bottomrule
\end{tabular}
\end{table}

\begin{table}[H]
\centering
\caption{router: per-seed semantic accuracy (\%).}
\small
\begin{tabular}{@{}lrrrrrrrc@{}}
\toprule
Seed & Baseline & step 20 & margin1 & margin3 & margin5 & final & $\Delta$ & $|\Delta| \le 5$\\
\midrule
0 & 93.75 & 85.62 & 78.12 & 59.38 & 61.88 & 64.38 & $-29.37$ & FAIL\\
1 & 93.75 & 87.50 & 86.88 & 61.25 & 66.88 & 76.88 & $-16.87$ & FAIL\\
2 & 93.75 & 86.25 & 83.75 & 61.25 & 60.62 & 58.13 & $-35.62$ & FAIL\\
3 & 93.75 & 83.75 & 77.50 & 65.00 & 61.25 & 63.12 & $-30.63$ & FAIL\\
4 & 93.75 & 82.50 & 80.62 & 54.37 & 69.38 & 61.88 & $-31.87$ & FAIL\\
5 & 93.75 & 84.38 & 82.50 & 44.38 & 51.88 & 48.75 & $-45.00$ & FAIL\\
6 & 93.75 & 85.62 & 83.75 & 61.25 & 64.38 & 69.38 & $-24.38$ & FAIL\\
7 & 93.75 & 85.62 & 79.38 & 62.50 & 64.38 & 66.25 & $-27.50$ & FAIL\\
\bottomrule
\end{tabular}
\end{table}

The per-seed $|\Delta| \le 5$ check is shown in the final column; item-level bootstrap 90\% CIs at the final checkpoint are reported per task in Table~\ref{tab:tost}.

Seeds 0 and 1 are deterministic re-runs of the pilot (\S\ref{sec:method}). Restricting to seeds 2--7: extraction mean $\Delta = -0.21$, json mean $\Delta = -3.33$, router mean $\Delta = -32.50$. All three verdicts are unchanged on the 6-seed subset (seed-level $t$ intervals with df $= 5$; item-level bootstrap as in the main analysis). Extraction remains equivalent: seed-level 90\% CI $[-2.18, +1.76]$, item-level $[-1.67, +1.35]$, and 6/6 seeds within $\pm5$. For json, equivalence remains not shown: its seed-level CI $[-4.29, -2.38]$ and 6/6 seeds are within $\pm5$, but the item-level CI $[-5.42, -1.46]$ crosses $-5$. Router remains not equivalent: seed-level CI $[-38.44, -26.56]$, 0/6 seeds within $\pm5$. The router failure-mode decomposition across the checkpoint trajectory is in main-text Table~\ref{tab:router-traj}; Figure~\ref{fig:traj} visualizes the same trajectory per seed.

\section{Scorer v2 Rules and Artifact Walkthrough}
\label{app:scorer}
\textbf{Scorer architecture.} Scorer v2 applies a deterministic four-tier classification to each generated output, evaluated in this order (\texttt{scorer\_v2.py}). (1)~\textbf{FORMAT\_BREAK}: the output fails the task's parse gate. For router, the output (after leading whitespace) must begin with one of the five valid labels as a delimited token (case-insensitive; trailing text allowed), so syntactically plausible but invalid labels such as invented UPPER\_SNAKE\_CASE pseudo-labels are FORMAT\_BREAK. For json and extraction, the output must begin with a JSON object (optionally code-fenced) that parses and contains the required keys. (2)~\textbf{PASS}: strict match to gold (field-wise exact match after whitespace and case normalization; dates normalized to YYYY-MM-DD). (3)~\textbf{SURFACE\_FORM}: no strict match, but a match after canonicalization (e.g., ``Item 1'' versus ``1''); only extraction has a canonicalization rule. (4)~\textbf{SEMANTIC\_ERROR}: no match after canonicalization (for router and json, any parsed output that fails the strict match). Semantic accuracy counts PASS $+$ SURFACE\_FORM as correct; FORMAT\_BREAK and SEMANTIC\_ERROR are failures.

\textbf{Canonicalization rules (extraction).} The section-identifier field is canonicalized before comparison. The numeric portion is extracted via regex \verb|\d+[A-Z]?| (``Item 7A'' $\to$ ``7A'', ``Item 1'' $\to$ ``1''). The alpha suffix is case-folded, and appended titles are ignored (``Item 7: Management's Discussion'' $\to$ ``7''). Ticker uses case-insensitive exact match; filing\_date is normalized to YYYY-MM-DD before exact match.

\textbf{The artifact: $0.550 \to 0.981$.} At the margin3 checkpoint, extraction accuracy under strict string matching is 0.550 (seed0); under scorer v2's canonicalization the same checkpoint scores 0.981. The 69 items ($69/72 = 96\%$) that flip from strict-FAIL to SURFACE\_FORM (semantically correct) are exclusively section-identifier surface-form variants. The model outputs ``1'' where gold says ``Item 1'', or ``7A'' where gold says ``Item 7A''. The semantic content is correct in every case, with no ticker or filing\_date errors. The 8-seed mean is $0.616 \to 0.988$. The register-shift rate (SURFACE\_FORM tier) peaks at margin ${\approx}3$ (mean 0.37 across 8 seeds), then subsides to 0.09 by final. This mirrors router's non-monotonic FORMAT\_BREAK trajectory. The artifact was identified in the pilot (seed 0 is bitwise identical in the pilot and confirmatory runs; pilot chronology, Note 3, committed 2026-06-11). The canonicalization rule was written and frozen in response, and the matrix evaluation was switched to scorer v2 before Note 2 was frozen (2026-06-18) and before the confirmatory run; the 8-seed figures above are from the confirmatory run. No rules were added or modified after observing its results.

\section{Pre-Registration Reconciliation and Falsification Chronology}
\label{app:reconcile}
\textbf{C.1 Secondary reconciliation.} Every pre-registered secondary from Note 2 (frozen 2026-06-18) is accounted for below.

\begin{table}[H]
\centering
\small
\caption{Pre-registered secondary reconciliation.}
\begin{tabular}{@{}clp{3.2cm}p{3.4cm}@{}}
\toprule
\# & Secondary & Status & Location\\
\midrule
1 & Trajectory-min per seed/task & Partly reported (router only) & \S\ref{sec:nonmono} (router minimum at margin3, 6/8)\\
2 & FORMAT\_BREAK trajectory & Reported & \S\ref{sec:nonmono} (Table~\ref{tab:router-traj})\\
3 & Held-out margin per category & Reported; computed under fp16 (Mac) and nf4 (Kaggle). \textbf{Falsified under both backends}: the prediction (style $>$ tool at every checkpoint) holds at 0/4 checkpoints under fp16 and 2/4 under nf4. Final-checkpoint per-pair ordering is backend-dependent. & \S\ref{sec:mechanism} (Tables~\ref{tab:heldout-fp16}, \ref{tab:heldout-nf4}, \ref{tab:heldout-nf4-perseed})\\
4 & Chosen/rejected reward decomposition & Reported (post-hoc; nf4 Kaggle, fp16 Mac) & \S\ref{sec:mechanism}, Fig.~\ref{fig:reward} (per-axis Table~\ref{tab:peraxis})\\
5 & Gold-logp trajectory $\times$ 3 tasks & Partly reported (endpoint $\Delta_{\text{gold}}$ only; per-checkpoint trajectory in Appendix~G.4 for KL only) & \S\ref{sec:gold} (Table~\ref{tab:gold})\\
6 & Probe statistic replication (paired) & Extraction 8/8 GOLD\_PROTECTED; json/router not computable (extraction-specific controls) & \S\ref{sec:gold} (Table~\ref{tab:protected})\\
7 & Register-shift tracking & Reported (SURFACE\_FORM rate) & \S\ref{sec:nonmono}\\
8 & KL-from-reference per checkpoint & Reported (post-hoc, fp16, 8 seeds) & \S\ref{sec:gold} (Appendix~\ref{app:gold})\\
\bottomrule
\end{tabular}
\end{table}

Secondaries \#4 and \#8 were not logged during training in the confirmatory run. \#4 was computed via a dedicated GPU inference run on saved adapters (Kaggle T4). \#8 was computed via Mac inference on saved adapters (fp16, proper token-level KL). \#3 was computed under fp16 (Mac) and nf4 (Kaggle T4) on saved adapters. All three use the same checkpoint adapters as the confirmatory run; no retraining was involved. The primary TOST analysis (\S\ref{sec:graded}) was unaffected.

\textbf{C.2 Pre-registration deviations.} \emph{Checkpoint mechanism:} the pre-registered design (Note 2) specified fixed-step checkpoints with margin as a post-hoc label; the implementation used margin-crossing triggers for intermediate checkpoints (margin1: steps 24--30, margin3: 84--98, margin5: 112--138). Step 20 and final (step 180) are fixed and identical across seeds, so the primary endpoint is unaffected. Cross-seed averages at intermediate checkpoints pool seeds at matched margin but different optimizer steps. \emph{Early-stop configuration:} the design specified \texttt{max\_opt\_steps=180} and \texttt{no\_early\_stop=True}. Both were enforced at runtime in \texttt{matrix\_kernel.py} and \texttt{run\_matrix.py}, and all 8 seeds terminated at exactly step 180. However, \texttt{train\_config.json} records module-level constants only and does not serialize these runtime parameters. This may create a misleading impression when inspecting saved artifacts. The execution matched the pre-registration; the config-file serialization did not. \emph{Seed overlap:} seeds 0 and 1 of the confirmatory set are bitwise re-runs of the pilot seeds (seed 0 fully; seed 1 through margin5). Six of eight confirmatory seeds were unseen at freeze time. The repository was first pushed publicly on 2026-10-18; the commit dates reflect local authorship and are not independently timestamped. \emph{Backend:} the pre-registration did not specify a quantization backend for any secondary. Secondary \#3 (held-out per-category margin) was computed under fp16 (Mac) and nf4 (Kaggle T4); the per-pair ordering is backend-dependent at the final checkpoint (\S\ref{sec:mechanism}).

\textbf{C.3 Falsification chronology.} \emph{Design phase (before the confirmatory run):} (a)~A causal-clamp experiment was designed but dropped as confounded. Clamping the margin also alters the KL penalty and total optimization pressure, making ``margin magnitude $\to$ capability loss'' untestable. (b)~Router was initially the sole primary endpoint. After pilot analysis revealed a non-monotonic transient, router's trajectory analysis became a secondary; router's final accuracy remained a primary TOST endpoint alongside json and extraction. (c)~A targeted-displacement hypothesis was reformulated as ``selective deflation.'' A preliminary probe had found that controls fell more than gold, indicating global drift rather than targeted displacement. (d)~Three exploration-phase numbers lacking on-disk artifacts were re-measured or dropped. The 80-item baselines were re-derived from the 160-item suite, the router baseline was corrected from 0.950 to 0.938, and gate floors were re-derived (extraction 0.006, router 0.200, json 0.0125). (e)~The suite was expanded from 80 to 160 items per task to improve TOST power, and frozen before the confirmatory run. (f)~The GOLD\_PROTECTED threshold ($X = 0.05$) was frozen before a preliminary 12-item probe existed. (g)~\textbf{Falsified:} the extraction collapse at margin ${\approx}3$. In the pilot, extraction accuracy at margin3 appeared to crash under the strict scorer (seed 0: 0.550); canonicalized scoring showed a surface-form register shift (0.981). The canonicalization rule was written and frozen in response, before the confirmatory run, and the strict-scorer measurement was discarded, not refined (Appendix~\ref{app:scorer}).

\emph{Confirmatory-run phase:} \textbf{Falsified:} (i)~the bounded-null headline (json and router failed the $\pm5$ TOST); (ii)~style margin $>$ tool\_calling on held-out. Item (ii) is falsified under both backends: the prediction (style $>$ tool at every checkpoint) holds at 0/4 checkpoints under fp16 and 2/4 under nf4. Final-checkpoint per-pair ordering is backend-dependent. \textbf{Refined:} margin-independent early drop became a damage window at margin ${\approx}1$--3 with two failure modes with distinct aggregate trajectories. ``Gold absolutely stable'' ($|\Delta_{\text{gold}}| \le 0.01$, nf4) became ``gold selectively protected but task-selectively deflates'' (fp16 revealed json $-1.04$, router $-1.47$, extraction $-0.15$).

\section{Suite Hygiene}
\label{app:hygiene}
Source files: \nolinkurl{data_eval/track1_{router,json,extraction}.jsonl}; leakage references \nolinkurl{kaggle_upload/training_data/dpo_pytorch.jsonl} (preference) and \nolinkurl{training_data/sft_pytorch/{train,valid,test}.jsonl} (SFT); degeneracy floors \nolinkurl{kaggle_upload/audit_eval.py:91}; gold self-consistency via \nolinkurl{kaggle_upload/scorer.py}.

\begin{table}[H]
\centering
\small
\caption{Suite hygiene properties.}
\begin{tabular}{@{}lccc@{}}
\toprule
Property & extraction & json & router\\
\midrule
$n$ items & 160 & 160 & 160\\
Unique prompts & 160 & 160 & 160\\
Duplicate prompts & 0 & 0 & 0\\
Verbatim leakage (eval $\cap$ preference) & 0 & 0 & 0\\
Verbatim leakage (eval $\cap$ SFT) & 0 & 0 & 0\\
Gold self-consistency (gold-as-output PASS) & 160/160 & 160/160 & 160/160\\
Degeneracy floor (always-one-answer) & ${\sim}0.006$ & 0.0125 & 0.20\\
\bottomrule
\end{tabular}
\end{table}

\textbf{Class balance.} \emph{router} (5 classes): COMPANY\_DEEP\_DIVE, FILING\_ANALYSIS, FILING\_DIFF, OUT\_OF\_SCOPE, SIMPLE\_QUERY $= 32$ each. \emph{json} (8 tool classes): analyst\_ratings, company\_metrics, company\_news, diff\_filing\_sections, get\_filing\_section, insider\_trades, search\_filings, stock\_price $= 20$ each. \emph{extraction} (6 section classes): Item 1 $=27$, Item 1A $=27$, Item 3 $=26$, Item 7 $=27$, Item 7A $=26$, Item 8 $=27$.

Zero verbatim leakage: 0/160 eval prompts appear as a preference-data \texttt{question} (exact-string match) for all three tasks, and 0/160 appear as an SFT user prompt (1{,}060 unique prompts across the SFT train/valid/test splits; exact-string and substring match). Gold self-consistency: 160/160 (each gold answer passes scorer v2; this establishes scorer compatibility, not independent factual verification). Zero duplicate prompts: 160 unique prompts / 160 items. Degeneracy floors (a constant always-one-answer policy): extraction $0.006$, router $0.20$ (5 balanced classes), json $0.0125$ ($2/160$). Extraction golds are unique (constant-output floor $1/160 \approx 0.006$); router has 5 balanced classes ($32/160 = 0.20$); json has 18 duplicate tool-call tuples ($2/160 = 0.0125$, maximum multiplicity 2).

\section{Training Configuration and Hash Verification}
\label{app:config}
\textbf{SFT checkpoint:} hash \texttt{29a3794b\ldots585f02}, verified identical across all 8 seed runs; adapter weights loaded from \texttt{adapters/sft\_pytorch/adapter\_model.safetensors}. The SFT adapter was trained with QLoRA (nf4, double quantization, fp16 compute) and the same LoRA configuration as below, for 4 epochs (lr $10^{-4}$, cosine schedule, 20 warmup steps, batch 1 $\times$ 8 gradient accumulation, max length 2048; \texttt{train\_sft\_kaggle.py}, saved \texttt{training\_args.bin}) on 862 training examples (108 validation, 108 test) in \texttt{training\_data/sft\_pytorch/}, generated by GPT-4o-mini. The 862 training examples comprise 365 analytical prose answers, 276 JSON tool calls, 130 router labels, and 91 answer-quality reviews (COMPLETE or NEEDS\_FOLLOW\_UP with a reason); none uses the extraction output format. DPO loads this adapter as trainable and continues training it; the frozen SFT adapter is the DPO reference. \textbf{Tokenizer and chat template:} the chat template is byte-identical across sources (SHA-256 hash comparison). For the tokenizer, tokenization is identical on all study inputs: the SFT adapter, DPO adapters, and base tokenizer configs differ in metadata (added-token count, pre-tokenizer regex) but produce identical token IDs on all 6{,}024 evaluation and preference strings, verified by direct token-ID comparison.

\textbf{LoRA configuration:} $r = 8$, $\alpha = 16$, dropout $= 0.05$; \texttt{target\_modules} $=$ [q\_proj, k\_proj, v\_proj, o\_proj] (explicitly configured, not peft default); bias $=$ none; peft version 0.19.1. Injected layers: 3, 7, 11, 15, 19, 23, 27, 31 (the 8 full-attention layers in Qwen3.5-4B's hybrid architecture; 1.57M trainable parameters, 0.034\% of 4.66B total).

\textbf{Training hyperparameters:} AdamW, lr $= 5\times10^{-5}$, constant schedule, weight\_decay $= 0.01$; gradient clipping 1.0; batch size 8 (gradient accumulation); $\beta$ (DPO) $= 0.5$; SFT retention weight $= 0.5$. Max optimizer steps are 180, enforced via a \texttt{matrix\_kernel.py} runtime parameter. Early stop is disabled (\texttt{no\_early\_stop=True} via \texttt{run\_matrix.py}). The divergence guard aborts if the loss is NaN or the smoothed margin exceeds 20.

\textbf{Config-file note:} \texttt{train\_config.json} records module-level constants and does not serialize the runtime parameters \texttt{max\_opt\_steps} and \texttt{no\_early\_stop}; runtime enforcement was applied in \texttt{matrix\_kernel.py} and \texttt{run\_matrix.py}, confirmed by source-code inspection. All 8 seeds terminated at exactly step 180. Seeds 0--4 and 5--7 were run in two separate Kaggle GPU quota periods. \textbf{Checkpoint adapters on disk:} 8 seeds $\times$ 5 checkpoints $= 40$ adapter directories, all verified present with complete \texttt{adapter\_model.safetensors} files.

\section{Reward Decomposition and Held-Out Margin Per-Seed}
\label{app:reward-data}
Source: \texttt{reward\_decomposition/summary.json} ($\beta{=}0.5$, reference $=$ sft\_baseline). $r = \text{policy\_logp} - \text{sft\_logp}$ (length-normalized); margin $= \beta\cdot(r_{\text{chosen}} - r_{\text{rejected}})$. The 8-seed-mean per-axis decomposition under both backends is in main-text Table~\ref{tab:peraxis}; per-seed nf4 values follow.

\begin{table}[H]
\centering
\small
\caption{F.1: $r_{\text{chosen}}$, $r_{\text{rejected}}$, margin per checkpoint, per seed (nf4).}
\label{tab:reward-perseed}
\begin{tabular}{@{}llrrr@{}}
\toprule
Seed & Ckpt & $r_{\text{chosen}}$ & $r_{\text{rejected}}$ & margin\\
\midrule
0 & margin1 & $-0.044$ & $-2.087$ & 1.021\\
0 & margin3 & $-0.094$ & $-5.569$ & 2.738\\
0 & margin5 & $-0.001$ & $-11.016$ & 5.508\\
0 & final & $0.051$ & $-14.518$ & 7.285\\
1 & margin1 & $-0.082$ & $-2.365$ & 1.141\\
1 & margin3 & $-0.060$ & $-6.380$ & 3.160\\
1 & margin5 & $0.006$ & $-10.146$ & 5.076\\
1 & final & $0.055$ & $-15.344$ & 7.699\\
2 & margin1 & $-0.039$ & $-2.075$ & 1.018\\
2 & margin3 & $-0.063$ & $-5.714$ & 2.825\\
2 & margin5 & $-0.033$ & $-10.861$ & 5.414\\
2 & final & $0.052$ & $-15.909$ & 7.981\\
3 & margin1 & $-0.045$ & $-2.024$ & 0.990\\
3 & margin3 & $-0.052$ & $-5.822$ & 2.885\\
3 & margin5 & $-0.033$ & $-9.320$ & 4.644\\
3 & final & $0.060$ & $-14.615$ & 7.337\\
4 & margin1 & $-0.054$ & $-2.285$ & 1.116\\
4 & margin3 & $-0.074$ & $-6.235$ & 3.080\\
4 & margin5 & $0.013$ & $-8.973$ & 4.493\\
4 & final & $0.054$ & $-12.302$ & 6.178\\
5 & margin1 & $-0.047$ & $-2.014$ & 0.984\\
5 & margin3 & $-0.100$ & $-6.196$ & 3.048\\
5 & margin5 & $0.018$ & $-9.233$ & 4.626\\
5 & final & $0.059$ & $-12.631$ & 6.345\\
6 & margin1 & $-0.051$ & $-2.005$ & 0.977\\
6 & margin3 & $-0.075$ & $-5.799$ & 2.862\\
6 & margin5 & $-0.023$ & $-11.535$ & 5.756\\
6 & final & $0.057$ & $-13.346$ & 6.701\\
7 & margin1 & $-0.051$ & $-2.052$ & 1.001\\
7 & margin3 & $-0.057$ & $-6.206$ & 3.074\\
7 & margin5 & $0.020$ & $-10.166$ & 5.093\\
7 & final & $0.032$ & $-14.890$ & 7.461\\
\bottomrule
\end{tabular}
\end{table}

\begin{table}[H]
\centering
\small
\caption{F.2: Per-axis margin at final checkpoint (style versus tool\_calling), per seed (nf4).}
\label{tab:peraxis-perseed}
\begin{tabular}{@{}lrrrrrr@{}}
\toprule
Seed & style $r_{\text{ch}}$ & style $r_{\text{rej}}$ & style margin & tool $r_{\text{ch}}$ & tool $r_{\text{rej}}$ & tool margin\\
\midrule
0 & $0.023$ & $-16.560$ & 8.291 & $0.113$ & $-10.005$ & 5.059\\
1 & $0.013$ & $-17.726$ & 8.870 & $0.147$ & $-10.077$ & 5.112\\
2 & $0.013$ & $-18.820$ & 9.417 & $0.139$ & $-9.475$ & 4.807\\
3 & $0.019$ & $-16.696$ & 8.358 & $0.149$ & $-10.016$ & 5.082\\
4 & $0.013$ & $-13.536$ & 6.774 & $0.147$ & $-9.572$ & 4.860\\
5 & $0.018$ & $-14.142$ & 7.080 & $0.149$ & $-9.292$ & 4.720\\
6 & $0.016$ & $-15.118$ & 7.567 & $0.148$ & $-9.428$ & 4.788\\
7 & $-0.019$ & $-17.206$ & 8.594 & $0.144$ & $-9.770$ & 4.957\\
\bottomrule
\end{tabular}
\end{table}

\begin{table}[H]
\centering
\small
\caption{F.3: Held-out margin by category per checkpoint, per seed (fp16; 98 pairs: 69 style, 29 tool\_calling; $\beta{=}0.5$). Under fp16, tool\_calling margin exceeds style margin at all four checkpoints for all 8 seeds.}
\label{tab:heldout-fp16}
\begin{tabular}{@{}llcc@{}}
\toprule
Seed & Ckpt & style margin & tool\_calling margin\\
\midrule
0 & margin1 & 0.650 & 1.301\\
0 & margin3 & 1.738 & 2.664\\
0 & margin5 & 2.480 & 3.253\\
0 & final & 2.501 & 4.034\\
1 & margin1 & 0.698 & 1.484\\
1 & margin3 & 1.843 & 3.069\\
1 & margin5 & 2.241 & 3.383\\
1 & final & 2.476 & 3.905\\
2 & margin1 & 0.615 & 1.361\\
2 & margin3 & 1.729 & 2.961\\
2 & margin5 & 2.498 & 3.285\\
2 & final & 1.904 & 3.709\\
3 & margin1 & 0.671 & 1.189\\
3 & margin3 & 1.987 & 2.743\\
3 & margin5 & 2.989 & 3.184\\
3 & final & 2.230 & 4.002\\
4 & margin1 & 0.721 & 1.411\\
4 & margin3 & 1.882 & 2.895\\
4 & margin5 & 2.044 & 3.270\\
4 & final & 2.115 & 3.856\\
5 & margin1 & 0.595 & 1.230\\
5 & margin3 & 1.977 & 2.782\\
5 & margin5 & 2.225 & 3.230\\
5 & final & 1.998 & 3.814\\
6 & margin1 & 0.680 & 0.952\\
6 & margin3 & 1.859 & 2.777\\
6 & margin5 & 2.568 & 3.297\\
6 & final & 1.998 & 3.739\\
7 & margin1 & 0.706 & 1.077\\
7 & margin3 & 2.055 & 2.913\\
7 & margin5 & 2.490 & 3.505\\
7 & final & 3.354 & 3.908\\
\bottomrule
\end{tabular}
\end{table}

\begin{table}[H]
\centering
\small
\caption{F.4: Held-out per-category margin under nf4 (8-seed mean; 98 pairs: 69 style, 29 tool\_calling). Tool\_calling leads through margin3; style overtakes by margin5 under axis-dependent nf4 inflation.}
\label{tab:heldout-nf4}
\begin{tabular}{@{}lcc@{}}
\toprule
Checkpoint & Style (nf4) & Tool (nf4)\\
\midrule
margin1 & 0.780 & 1.466\\
margin3 & 2.620 & 3.426\\
margin5 & 5.435 & 4.020\\
final & 7.919 & 4.753\\
\bottomrule
\end{tabular}
\end{table}

\begin{table}[H]
\centering
\small
\caption{F.5: Held-out margin by category per checkpoint, per seed (nf4; 98 pairs: 69 style, 29 tool\_calling; $\beta{=}0.5$; source \texttt{kaggle\_result/val\_nf4\_98.json}). Tool\_calling margin exceeds style margin at margin1 and margin3 for all 8 seeds; style exceeds tool\_calling at margin5 and final for all 8 seeds.}
\label{tab:heldout-nf4-perseed}
\begin{tabular}{@{}lcccccccc@{}}
\toprule
 & \multicolumn{2}{c}{margin1} & \multicolumn{2}{c}{margin3} & \multicolumn{2}{c}{margin5} & \multicolumn{2}{c}{final}\\
Seed & style & tool & style & tool & style & tool & style & tool\\
\midrule
0 & 0.757 & 1.518 & 2.420 & 3.118 & 6.107 & 3.951 & 8.027 & 4.923\\
1 & 0.831 & 1.603 & 2.757 & 3.734 & 5.384 & 4.167 & 8.677 & 4.943\\
2 & 0.719 & 1.544 & 2.361 & 3.466 & 5.943 & 3.912 & 9.166 & 4.596\\
3 & 0.764 & 1.399 & 2.558 & 3.304 & 4.984 & 3.798 & 8.096 & 4.914\\
4 & 0.829 & 1.643 & 2.701 & 3.510 & 4.501 & 3.974 & 6.728 & 4.696\\
5 & 0.702 & 1.543 & 2.782 & 3.401 & 4.897 & 3.981 & 6.969 & 4.527\\
6 & 0.820 & 1.201 & 2.545 & 3.381 & 6.277 & 4.187 & 7.315 & 4.667\\
7 & 0.816 & 1.275 & 2.832 & 3.495 & 5.389 & 4.189 & 8.372 & 4.756\\
\bottomrule
\end{tabular}
\end{table}

\begin{table}[H]
\centering
\small
\caption{F.6: fp16 train reward decomposition, pooled per checkpoint (8-seed mean; 549 train pairs: 378 style, 171 tool\_calling; $\beta{=}0.5$, reference $=$ sft\_baseline). Chosen dips transiently at margin3 ($-0.173$) before recovering; rejected falls monotonically.}
\label{tab:fp16-train-pooled}
\begin{tabular}{@{}lccc@{}}
\toprule
Checkpoint & $r_{\text{chosen}}$ & $r_{\text{rejected}}$ & margin\\
\midrule
margin1 & $-0.045$ & $-1.785$ & 0.870\\
margin3 & $-0.173$ & $-4.661$ & 2.244\\
margin5 & $-0.126$ & $-5.743$ & 2.809\\
final & $-0.028$ & $-5.928$ & 2.950\\
\bottomrule
\end{tabular}
\end{table}

\section{Gold-Logp Per-Seed Data, nf4 Cross-Check, and KL Trajectory}
\label{app:gold}
\textbf{G.1 fp16 same-backend $\Delta_{\text{gold}}$ (per-seed).} Source: \texttt{gold\_3task\_summary.json} (fp16\_mac, reference $=$ sft\_baseline). $\Delta_{\text{gold}} = \text{gold(final)} - \text{gold(sft)}$, per-token length-normalized. Bracketed values are per-seed percentile bootstrap 95\% CIs over the 160 per-item differences (10{,}000 resamples); the across-seed CIs below bootstrap the eight per-seed means.

\begin{table}[H]
\centering
\small
\caption{G.1: per-seed fp16 $\Delta_{\text{gold}}$ with 95\% CI.}
\begin{tabular}{@{}lccc@{}}
\toprule
Seed & extraction & json & router\\
\midrule
0 & $-0.102\ [-0.108,-0.095]$ & $-1.505\ [-1.581,-1.430]$ & $-1.259\ [-1.645,-0.902]$\\
1 & $-0.226\ [-0.232,-0.220]$ & $-1.177\ [-1.281,-1.074]$ & $-1.616\ [-2.029,-1.226]$\\
2 & $-0.147\ [-0.153,-0.141]$ & $-1.001\ [-1.078,-0.925]$ & $-1.458\ [-1.903,-1.044]$\\
3 & $-0.102\ [-0.108,-0.096]$ & $-0.957\ [-1.057,-0.858]$ & $-1.305\ [-1.723,-0.919]$\\
4 & $-0.150\ [-0.157,-0.143]$ & $-0.906\ [-1.001,-0.812]$ & $-1.915\ [-2.406,-1.451]$\\
5 & $-0.194\ [-0.200,-0.188]$ & $-0.761\ [-0.854,-0.671]$ & $-1.307\ [-1.692,-0.939]$\\
6 & $-0.138\ [-0.145,-0.131]$ & $-1.175\ [-1.252,-1.101]$ & $-1.377\ [-1.749,-1.028]$\\
7 & $-0.168\ [-0.177,-0.160]$ & $-0.803\ [-0.904,-0.703]$ & $-1.539\ [-1.999,-1.109]$\\
\bottomrule
\end{tabular}
\end{table}

Across-seed means (8 seeds): extraction $-0.153\ [-0.182,-0.126]$, json $-1.036\ [-1.204,-0.891]$, router $-1.472\ [-1.624,-1.345]$.

\textbf{G.2 nf4 cross-check.} nf4 $\Delta_{\text{gold}}$ is computed as gold\_logp(final) $-$ gold\_logp(step 20), using the records of the confirmatory run; no nf4 SFT baseline gold-logp was recorded, so step 20 is the earliest available reference.

\begin{table}[H]
\centering
\small
\caption{G.2: nf4 $\Delta_{\text{gold}}$ (step 20$\to$final; nats/token), 8-seed mean. 95\% CI: $t$-interval across the eight per-seed means (df $= 7$).}
\begin{tabular}{@{}lccc@{}}
\toprule
Task & $\Delta_{\text{gold}}$ (nf4) & 95\% CI & $\Delta_{\text{gold}}$ (fp16) for comparison\\
\midrule
extraction & $-0.008$ & $[-0.014,-0.003]$ & $-0.15$\\
json & $-0.22$ & $[-0.34,-0.11]$ & $-1.04$\\
router & $-1.59$ & $[-1.99,-1.19]$ & $-1.47$\\
\bottomrule
\end{tabular}
\end{table}

\begin{table}[H]
\centering
\small
\caption{G.2: per-seed nf4 $\Delta_{\text{gold}}$ (step 20$\to$final).}
\begin{tabular}{@{}lccc@{}}
\toprule
Seed & extraction & json & router\\
\midrule
0 & $-0.0089$ & $-0.1495$ & $-1.6280$\\
1 & $-0.0134$ & $-0.5054$ & $-1.6686$\\
2 & $-0.0089$ & $-0.1291$ & $-1.0105$\\
3 & $-0.0004$ & $-0.3407$ & $-1.7108$\\
4 & $+0.0002$ & $-0.1034$ & $-1.9370$\\
5 & $-0.0164$ & $-0.1119$ & $-2.4182$\\
6 & $-0.0047$ & $-0.2077$ & $-0.9808$\\
7 & $-0.0141$ & $-0.2485$ & $-1.3311$\\
\bottomrule
\end{tabular}
\end{table}

The task-selective ordering (extraction $\ll$ json $<$ router) is identical under both backends, with non-overlapping CIs. Absolute magnitudes diverge. The comparison changes both the backend and the baseline (fp16 uses the pure SFT adapter; nf4 uses step 20, which has absorbed 20 DPO steps), so the divergence cannot be attributed to quantization alone. The divergence is itself task-selective: extraction's deflation nearly vanishes under nf4 (from $-0.15$ to $-0.008$), json compresses ${\sim}5\times$, while router's magnitude is essentially unchanged.

\textbf{G.3 GOLD\_PROTECTED paired statistic (fp16, extraction, per-seed).} Source: \nolinkurl{gold_controls_summary.json} (fp16 unquantized, paired stat $\Delta_{\text{gold}}-\Delta_{\text{mean}(A,B)}$, reference $=$ sft\_baseline, $X{=}0.05$). GOLD\_PROTECTED verdict when stat $\ge X$ and the 95\% bootstrap CI (over the 160 per-item paired differences) excludes zero.

\begin{table}[H]
\centering
\small
\caption{G.3: per-seed GOLD\_PROTECTED statistic.}
\begin{tabular}{@{}lccccc@{}}
\toprule
Seed & stat & 95\% CI & $\Delta_{\text{gold}}$ & $\Delta_{\text{ctrl}}$ (mean A,B) & Verdict\\
\midrule
0 & 0.2561 & $[0.2480,0.2644]$ & $-0.1016$ & $-0.3576$ & GOLD\_PROTECTED\\
1 & 0.2425 & $[0.2348,0.2502]$ & $-0.2262$ & $-0.4687$ & GOLD\_PROTECTED\\
2 & 0.2125 & $[0.2053,0.2198]$ & $-0.1470$ & $-0.3595$ & GOLD\_PROTECTED\\
3 & 0.2389 & $[0.2316,0.2462]$ & $-0.1022$ & $-0.3411$ & GOLD\_PROTECTED\\
4 & 0.2953 & $[0.2870,0.3038]$ & $-0.1502$ & $-0.4455$ & GOLD\_PROTECTED\\
5 & 0.2172 & $[0.2108,0.2236]$ & $-0.1938$ & $-0.4110$ & GOLD\_PROTECTED\\
6 & 0.2477 & $[0.2403,0.2553]$ & $-0.1379$ & $-0.3856$ & GOLD\_PROTECTED\\
7 & 0.2666 & $[0.2574,0.2756]$ & $-0.1684$ & $-0.4350$ & GOLD\_PROTECTED\\
\bottomrule
\end{tabular}
\end{table}

Aggregate: 8/8 seeds GOLD\_PROTECTED.

\textbf{G.4 KL(policy $\|$ reference) trajectory.}

\begin{table}[H]
\centering
\small
\caption{G.4: KL trajectory (extraction gold, fp16, 8-seed mean $\pm$ std).}
\begin{tabular}{@{}lcc@{}}
\toprule
Checkpoint & Mean KL & Std\\
\midrule
margin1 & 0.022 & 0.005\\
margin3 & 0.057 & 0.017\\
margin5 & 0.083 & 0.014\\
final & 0.082 & 0.008\\
\bottomrule
\end{tabular}
\end{table}

KL rises monotonically through margin5 then plateaus ($0.083 \to 0.082$), confirming mild policy drift (all values $<0.1$ nats/token) that saturates as margin continues to grow from ${\sim}5$ to ${\sim}7$. KL is computed on extraction gold completions only; drift on other tasks may be larger. Per-seed KL at final: seed0 $=0.071$, seed1 $=0.089$, seed2 $=0.081$, seed3 $=0.075$, seed4 $=0.077$, seed5 $=0.092$, seed6 $=0.077$, seed7 $=0.093$.

\section{Validation Margin Trajectories}
\label{app:valmargin}
Source: \texttt{track1\_dpo\_seed*\_margin.jsonl}. \texttt{val\_margin} is logged only at the evaluation cadence (\texttt{eval\_every} $= 50 \to$ steps 50, 100, 150); it is not recorded at the checkpoint steps. The table reports every logged val\_margin step.

\begin{table}[H]
\centering
\small
\caption{Validation margin at logged steps, with intermediate checkpoint steps for reference.}
\begin{tabular}{@{}lrrrrrr@{}}
\toprule
Seed & val@50 & val@100 & val@150 & (margin1 step) & (margin3 step) & (margin5 step)\\
\midrule
0 & 1.701 & 3.534 & 5.752 & 27 & 84 & 134\\
1 & 1.648 & 3.091 & 5.904 & 29 & 93 & 131\\
2 & 1.685 & 3.618 & 5.746 & 26 & 87 & 124\\
3 & 1.660 & 3.506 & 7.397 & 28 & 90 & 112\\
4 & 1.701 & 3.483 & 6.449 & 30 & 91 & 130\\
5 & 1.804 & 3.502 & 4.968 & 24 & 87 & 135\\
6 & 1.743 & 3.522 & 6.709 & 27 & 88 & 138\\
7 & 1.528 & 3.288 & 5.768 & 26 & 98 & 135\\
\bottomrule
\end{tabular}
\end{table}

The 8-seed mean val\_margin at step 150 is 6.087 (rounds to 6.1).

\textbf{Val-subset note.} The training-time \texttt{val\_margin} logged in the margin files is computed over a fixed 20-pair subset (15 style, 5 tool\_calling) of the 98 held-out pairs under the nf4 training backend, not the full 98-pair set. Post-hoc fp16 recomputation on the same 20 pairs yields a mean of 2.987 at the final checkpoint, versus 7.519 under nf4: a $2.52\times$ backend ratio. The logged \texttt{val\_margin} values are therefore not comparable to the fp16 held-out margins in Table~\ref{tab:heldout-fp16}.

\end{document}